\documentclass[10pt]{article}

\usepackage{PRIMEarxiv}

\usepackage[utf8]{inputenc} 
\usepackage[T1]{fontenc}    
\usepackage{hyperref}       
\usepackage{url}            
\usepackage{booktabs}       
\usepackage{amsfonts}       
\usepackage{nicefrac}       
\usepackage{enumitem}
\usepackage{lipsum}
\usepackage{multirow}
\usepackage[linesnumbered,ruled,vlined]{algorithm2e}
\SetKwInput{Input}{Input}   
\SetKwInput{Output}{Output} 
\usepackage{amsmath, amssymb, mathrsfs}
\usepackage{tabularx}
\usepackage{setspace}
\usepackage{subcaption}
\usepackage{xcolor}         
\usepackage{fancyhdr}       
\usepackage{graphicx}       
\usepackage[backend=biber, style=numeric, sorting=none, sortcites=true]{biblatex}
\graphicspath{{media/}}     
\usepackage{bm}

\DeclareMathOperator*{\argmin}{argmin} 
\DeclareMathOperator*{\argmax}{argmax} 
\title{Superpixel-based and Spatially-regularized Diffusion Learning for Unsupervised Hyperspectral Image Clustering
}

\author{
Kangning Cui\thanks{Also affiliated with the Hong Kong Centre for Cerebro-Cardiovascular Health Engineering, 19W, Science Park, Hong Kong} $^{,}$\thanks{{Corresponding author. Email: kangnicui2-c@my.cityu.edu.hk}}\\
Department of Mathematics\\
City University of Hong Kong \\
Kowloon, Hong Kong \\
\And
Ruoning Li \\
Department of Mathematics \\
City University of Hong Kong \\
Kowloon, Hong Kong \\
\And
Sam L. Polk \\
Department of Mathematics \\
Tufts University \\
Medford, MA, USA \\
\And
Yinyi Lin \\
Department of Geography \\
The University of Hong Kong \\
Pokfulam, Hong Kong \\
\And
Hongsheng Zhang \\
Department of Geography \\
The University of Hong Kong \\
Pokfulam, Hong Kong \\
\And
James M. Murphy \\
Department of Mathematics \\
Tufts University \\
Medford, MA, USA \\
\And
Robert J. Plemmons \\
Department of Computer Science \\
Wake Forest University \\
Winston-Salem, NC, USA \\
\And
Raymond H. Chan\footnotemark[1] \\
Department of Mathematics \\
City University of Hong Kong \\
Kowloon, Hong Kong\\
}

\begin{document}
\maketitle

\let\thefootnote\relax
\footnotetext{Code: \href{https://github.com/ckn3/S2DL}{github.com/ckn3/S2DL}}

\onehalfspacing
\begin{abstract}

Hyperspectral images (HSIs) provide exceptional spatial and spectral resolution of a scene, enabling a wide range of applications in remote sensing. However, the high dimensionality, presence of noise and outliers, and the need for precise labels of HSIs present significant challenges to the analysis of HSIs, motivating the development of performant HSI clustering algorithms. In this paper, we propose a novel unsupervised HSI clustering algorithm---Superpixel-based and Spatially-regularized Diffusion Learning (S$^2$DL)---which addresses these challenges by incorporating rich spatial information encoded in HSIs into a diffusion geometry-based clustering procedure. S$^2$DL employs the Entropy Rate Superpixel (ERS) segmentation technique to partition an image into superpixels. From these superpixels, a spatially-regularized diffusion graph is constructed using the most representative high-density pixels, reducing computational burden while maintaining an accurate perspective of the image. Cluster modes (points serving as exemplars for underlying cluster structure) are identified as the highest-density pixels farthest in diffusion distance (calculated using the aforementioned graph) from other highest-density pixels. These cluster modes are used to guide the labeling of the remaining representative pixels from ERS superpixels. Finally, majority voting is applied to the labels assigned within each superpixel to propagate labels to the rest of the image.
This spatial-spectral approach simultaneously simplifies graph construction, reduces computational cost, and improves clustering performance. The performance of S$^2$DL is illustrated with extensive numerical experiments on three publicly available, real-world HSIs: Indian Pines, Salinas, and Salinas A. In addition, we apply S$^2$DL to the important real-world problem of landscape-scale, unsupervised mangrove species mapping using a Gaofen-5 HSI generated over the Mai Po Nature Reserve, Hong Kong. The success of S$^2$DL in these diverse numerical experiments indicates its efficacy on a wide range of important unsupervised remote sensing analysis tasks. 

\end{abstract}

\keywords{Diffusion Geometry \and Superpixel Segmentation \and Spatial Regularization \and Hyperspectral Image Clustering \and Species Mapping}



\section{Introduction}
Hyperspectral images (HSIs) encode reflectance across a broad spectrum of wavelengths in the visual and infrared light spectra, storing a rich characterization of large spatial regions with high spectral resolution. These images can be obtained through various platforms, such as airplanes, drones, or orbital spectrometers~\cite{bioucas2013hyperspectral, plaza2009recent}. HSI data have been shown to be useful in a wide range of high-impact applications; e.g., the identification of land use and land cover~\cite{chen2014deep, zhai2021hyperspectral, li2019deep, polk2022unsupervised, li2022classification, liu2019review}, the unmixing of spectral signatures~\cite{borsoi2021spectral, bioucas2012hyperspectral, heylen2014review, cui2021unsupervised}, and the fusion of images from different sources and modalities~\cite{yokoya2017hyperspectral, loncan2015hyperspectral, liu2018deep, camalan2022detecting}. These applications often leverage machine learning and deep learning approaches to exploit the rich information stored within HSIs. However, the need for specialized knowledge and extensive fieldwork for expert annotations makes HSI labeling a resource-intensive and costly task, motivating the development of unsupervised clustering algorithms for analyzing HSIs~\cite{bioucas2013hyperspectral, li2019deep, zhao2021superpixel}. Nevertheless, even HSI clustering algorithms, which rely upon no expert annotations or ground truth labels for image segmentations, face at least two key challenges in practice~\cite{ding2022self, zhang2019hyperspectral, cai2020graph, peng2023graph}.

The first challenge for HSI clustering is the sheer volume of pixels in an image that the algorithm must analyze. HSIs being large datasets is a challenge independent to their inherent high-dimensionality, which stems from the high spectral resolution obtained using advanced hyperspectral spectrometers to generate high-quality HSI data. Large images are especially challenging for graph-based HSI clustering methods, which construct a graph to represent pixel affinities and can scale quadratically, if not worse, with the number of pixels if the graph is not constructed carefully~\cite{bioucas2013hyperspectral, zhai2021hyperspectral, huang2021hybrid, zhai2016new}. Several techniques have been proposed to mitigate the computational demands of the graph construction process. For instance, anchor-based graph approaches reduce the number of graph nodes by selecting representative anchor points, while the Nystr{\"o}m extension approximates the graph's eigendecomposition to streamline later analysis~\cite{wang2017fast, zhao2019fast, wang2022spatial}. However, the selection of anchors, often through random or $K$-means methods, can misrepresent the dataset, neglect spatial information, and be noise-sensitive~\cite{chen2023spectral}. Addressing the computational challenges, it's important to recognize that HSIs typically possess a low intrinsic dimensionality because of the natural relationships between spectral bands, which can often be captured in a manifold coordinate system~\cite{bachmann2005exploiting, murphy2018unsupervised, coifman2006diffusion}. This reduced intrinsic dimensionality motivates the downsampling of pixels, as a carefully selected subset of the HSI is likely to contain sufficient information to approximate the low-dimensional geometry intrinsic to an HSI~\cite{bachmann2005exploiting}.

A second major challenge to HSI clustering algorithms is the presence of noise and outliers in HSIs, which often results in a significant reduction in performance on important unsupervised material classification problems due to internal and external factors. These factors include sensor noise, atmospheric effects, and spectral variability, resulting in a noisy characterization of a scene in which it can be difficult to differentiate materials in an unsupervised setting~\cite{rasti2018noise, cai2023large}. A further complication is the high intra-class spectral variability often observed in different locations of an HSI resulting from variance in illumination conditions and viewing angles~\cite{borsoi2021spectral}. Integrating spatial information with spectral information can help mitigate the effects of spectral variability. This allows the algorithm to use contextual information from spatially nearby pixels, which often belong to the same class and share similar spectral characteristics~\cite{fauvel2012advances, jiang2018superpca, li2015efficient}. 

In this paper, we propose Superpixel-based and Spatially-regularized Diffusion Learning (S$^2$DL): a superpixel-based Diffusion Learning approach to unsupervised clustering of large HSIs. Our approach first employs Entropy Rate Superpixel (ERS) segmentation to partition the image into spatial regions of similarly-expressive pixels. ERS segmentation assumes local homogeneity and is meant to reduce the effect of spatial-spectral variability within each superpixel. S$^2$DL uses a kernel density estimate (KDE) to identify a small number of most-representative pixels from each superpixel for use in the construction of a spatially-regularized k-nearest neighbor (kNN) graph~\cite{murphy2018unsupervised, polk2021multiscale, murphy2019spectral, murphy2020spatially}. Importantly, the graph used in S$^2$DL explicitly incorporates spatial information into graph construction by allowing edges only between pixels and their nearest neighbors within a spatial radius. S$^2$DL locates and assigns unique labels to single pixels from each cluster to serve as cluster exemplars: highest density pixels farthest in spatially-regularized diffusion distances~\cite{murphy2018unsupervised, polk2021multiscale, murphy2019spectral, murphy2020spatially} from other high-density pixels. S$^2$DL then propagates labels across the graph using a local backbone (LBB) spread and diffusion-based label propagation. After the clustering of pixels representative of superpixels, majority voting is performed within each superpixel, ensuring spatial homogeneity of cluster assignments. As will become clear in extensive numerical experiments showing its efficacy, S$^2$DL's procedure results in a substantial decrease in computational complexity while at the same time mitigating the effects of noise and outliers.  Thus, S$^2$DL is a highly efficient and accurate approach to unsupervised HSI clustering. 

This article is organized as follows. In Section \ref{sec:background}, we discuss related work on HSI clustering, diffusion geometry, and spatial-spectral HSI clustering algorithms while overviewing terminology and background. Section 3  introduces S$^2$DL and motivates its approach to spatial-spectral clustering of HSIs. Section 4 shows the efficacy of S$^2$DL through extensive numerical experiments comparing S$^2$DL with classical and state-of-the-art unsupervised approaches on three benchmark HSI datasets: Indian Pines, Salinas and Salinas A. Additionally, this section showcases the application of S$^2$DL to a Gaofen-5 sensor-collected HSI dataset for unsupervised mapping mangrove species in Mai Po Nature Reserve, Hong Kong~\cite{wan2020gf}. Section 5 concludes and discusses future work.

\section{Related Works}
\label{sec:background}

\subsection{Overview of HSI Clustering Techniques} \label{sec:clustering}

Clustering is an unsupervised learning technique that groups similar objects or data points without the need for ground truth labels or expert annotations~\cite{friedman2001elements}. We denote pixels in an HSI as a set $X=\{x_i\}_{i=1}^N \subset~\mathbb{R}^B$, where each $x_i$ denotes the spectral signature of a pixel in the image, and $B$ represents the number of spectral bands. HSI clustering algorithms partition pixels into a clustering $\{X_k\}_{k=1}^K$ (where each $X_k$ is a cluster) such that pixels from the same cluster are ``similar'' (possibly due to a similar material constitution), while pixels from different clusters are ``dissimilar'' (containing different materials)~\cite{polk2023diffusion}. The specific notion of similarity used varies widely across the many clustering algorithms in the literature.

Traditional clustering methods---e.g., $K$-means~\cite{kmeans}, Gaussian Mixture Model~\cite{mclachlan2019finite}, and Density-Based Spatial Clustering of Applications with Noise~\cite{ester1996density}---have been employed for HSI clustering~\cite{zhai2021hyperspectral}. However, these methods often encounter challenges due to the presence of spectrally-mixed pixels and noise inherent in HSI data~\cite{polk2023diffusion, murphy2018unsupervised, rasti2018noise}; e.g., sensitivity to initialization, assumptions on the distribution of data~\cite{murphy2022multiscale, murphy2018unsupervised, maggioni2019learning, bachmann2005exploiting}, and sensitivity to noise~\cite{zhai2021hyperspectral}. Density peak clustering (DPC)~\cite{rodriguez2014clustering} was introduced to mitigate some of these distribution assumption errors (specifically, that of uniform density across clusters). DPC locates $K$ points in a dataset to serve as cluster modes: exemplars for latent underlying cluster structure. DPC cluster modes are defined as the highest-density points farthest in Euclidean distance from other high-density points. These cluster modes are assigned unique labels, which are propagated across the dataset by assigning (in order of decreasing density) each point the label of its $\ell^2$-nearest neighbor of higher density that is already labeled. However, the use of Euclidean distances to make pairwise comparisons between HSI pixels  in DPC~\cite{rodriguez2014clustering} and other traditional clustering algorithms~\cite{friedman2001elements} has been shown to reduce clustering quality on datasets with nonlinear decision boundaries between latent clusters~\cite{polk2023diffusion, bachmann2005exploiting, murphy2019spectral}, or due to the ``curse of dimensionality,'' that all pixels tend to appear roughly equidistant from one another in high-dimensional space~\cite{zhai2021hyperspectral, murphy2018unsupervised}. Finally, these methods often cluster individual pixels, agnostic to rich spatial information present in HSI data~\cite{bioucas2013hyperspectral, zhai2021hyperspectral, borsoi2021spectral, fauvel2012advances, murphy2019spectral}.

Deep clustering algorithms have been investigated for use on HSIs due to their ability to capture nonlinear decision boundaries and learn discriminative features for material classification~\cite{zhai2021hyperspectral, li2019deep, camalan2022detecting, min2018survey, cai2021large, cai2022superpixel, ding2022unsupervised, li2021self}. The vast deep clustering techniques in the literature range from contrastive learning-based~\cite{cai2021large, cai2022superpixel} to graph-based approaches~\cite{ding2022unsupervised, li2021self}. Notably, the inclusion of superpixels in deep HSI analysis has been shown to be beneficial for both computational efficiency and semantic consistency~\cite{cai2022superpixel, bandyopadhyay2022tree}. Despite their success on a wide range of HSI clustering problems, deep clustering algorithms are often highly sensitive to noise and perturbations in the data~\cite{nguyen2015deep, szegedy2013intriguing} and recent research has indicated that their success may be attributable to preprocessing steps rather than the learning capabilities of the network itself~\cite{haeffele2020critique}. Moreover, despite their highly accurate recovery of latent material structure in some applications, many deep HSI clustering approaches face limitations such as high computational complexity of training and the need for large training sets~\cite{min2018survey, cai2022superpixel, li2021self}.

\subsection{Diffusion Learning for HSI}
\label{sec:dlHSI}

Diffusion Learning has emerged as a highly effective approach to extracting and utilizing the inherent geometric structure contained within HSI data in an unsupervised clustering framework~\cite{polk2023diffusion, murphy2019spectral, coifman2006diffusion, murphy2018unsupervised, maggioni2019learning}. Diffusion Learning interprets HSI pixels as nodes in an undirected, weighted graph, the edges between which indicate pairwise similarity between pixels~\cite{ng2001spectral, murphy2018unsupervised}. This graph can be stored in a sparse adjacency matrix $\mathbf{W}\in\mathbb{R}^{N\times N}$, where $\mathbf{W}_{ij}=1$ if the pixel $x_j$ is one of the $k_{n}$ $\ell^2$-nearest neighbors (where $\ell^2$ denotes the Euclidean distance) of the pixel $x_i$ in $X$, and $\mathbf{W}_{ij}=0$ otherwise. Diffusion Learning relies on the data-dependent diffusion distance metric to make pairwise comparisons between pixels in the HSI~\cite{murphy2018unsupervised, coifman2006diffusion}, enabling highly accurate extraction of latent nonlinear structure in HSIs~\cite{murphy2018unsupervised, polk2021multiscale, murphy2022multiscale, polk2023diffusion, polk2022unsupervised, polk2022active}. 

Diffusion distances are calculated by considering a Markov diffusion process on the graph underlying $\mathbf{W}$~\cite{coifman2006diffusion}. The transition matrix for this diffusion process can be calculated directly from $\mathbf{W}$: $\mathbf{P} = \mathbf{D}^{-1}\mathbf{W}$, where $\mathbf{D}$ is the diagonal degree matrix with $\mathbf{D}_{ii} = \sum_{j=1}^N \mathbf{W}_{ij}$. 
Provided the graph underlying $\mathbf{P}$ is irreducible and aperiodic, $\mathbf{P}$ will possess a unique stationary distribution $\pi\in\mathbb{R}^{1\times N}$ such that $\pi \mathbf{P}=\pi$. 
The diffusion distance at a time $t\geq 0$ between any two pixels in the HSI $x_i,x_j\in X$~\cite{coifman2006diffusion, maggioni2019learning, murphy2022multiscale, polk2021multiscale} is defined by 
\begin{equation}
    D_t(x_i, x_j) = \sqrt{\sum_{k=1}^N [(\mathbf{P}^t)_{ik}-(\mathbf{P}^t)_{jk}]^2/\pi_k }.
\end{equation}
Diffusion distances have a natural relationship to the clustering problem~\cite{polk2023diffusion, coifman2006diffusion, maggioni2019learning}. Indeed, one may expect many high-weight length-$t$ paths between pixels sampled from the same latent cluster but very few such paths between pixels sampled from different clusters, resulting in intra-cluster diffusion distances that are small relative to inter-cluster diffusion distances~\cite{polk2023diffusion, coifman2006diffusion, maggioni2019learning}. The diffusion time parameter $t$ governs the scale of structure considered by diffusion distances, with smaller $t$ enabling retrieval of small-scale local structure in the image and larger $t$ retrieving global structure~\cite{polk2021multiscale, murphy2022multiscale, coifman2006diffusion}. 

Importantly, the eigendecomposition of $\mathbf{P}$ can be used for the efficient computation of diffusion distances. Indeed, given the eigenvalue-eigenvector pairs $\{(\lambda_k, \psi_k)\}_{k=1}^N$ of the transition matrix $\mathbf{P}$, it can be shown~\cite{coifman2006diffusion} that 
\begin{equation}
    D_t(x_i,x_j) = \sqrt{\sum_{k=1}^N|\lambda_k|^{2t}[(\psi_k)_i -(\psi_k)_j]^2}
\end{equation}
for $t\geq 0$ and $x_i,x_j\in X$. Crucially, under the assumptions of irreducibility and aperiodicity, $|\lambda_k| < 1$ for $k > 1$. This implies that, for sufficiently large $t$, diffusion distances can be accurately approximated using only the few eigenvectors $\psi_k$ corresponding to the largest $|\lambda_k|$~\cite{coifman2006diffusion, maggioni2019learning, murphy2022multiscale}. 

Employing diffusion geometry for HSI clustering yields significant advantages. Diffusion distances effectively counter the ``curse of dimensionality'' inherent to high-dimensional datasets like HSIs by extracting an intrinsic lower-dimensional, nonlinear representation of pixels, simultaneously reducing sensitivity to noise and redundant information~\cite{murphy2018unsupervised, maggioni2019learning, bachmann2005exploiting}. Consequently, these methods provide a robust approach to extracting latent geometric structure hidden in high-dimensional HSI data~\cite{murphy2018unsupervised, polk2021multiscale, polk2023diffusion} and have strong performance guarantees on clustering recovery across wide classes of data types~\cite{maggioni2019learning,murphy2022multiscale}. In addition, as will be discussed soon, there exist spatial-spectral variants of Diffusion Learning, achieved by constructing a spatially-regularized graph or assigning labels via spatial window constraints, enhancing these algorithms' sensitivity to the spatial context of each HSI pixel and therefore improving the quality of derived clusters~\cite{murphy2019spectral, polk2021multiscale, cui2022unsupervised, murphy2020spatially}. 

\subsection{Spatial-Spectral Analysis of HSI}\label{sec: spectral-spatial}

Incorporating the spatial structure of an HSI into an HSI clustering algorithm is essential to achieving high-quality partitions and mitigating the effects of noise and spectral variability in HSIs. This is due to the tendency of nearby pixels to exhibit similar spectral properties~\cite{zhai2021hyperspectral, cui2022unsupervised, chen2023spectral, zhao2019fast, min2018survey, cai2022superpixel, ding2022unsupervised}. Spatially-regularized graphs offer a robust framework for embedding spatial context into graph-based HSI clustering algorithms like Diffusion Learning~\cite{murphy2019spectral, polk2021multiscale}. Traditional graph-based methods that focus solely on spectral information and are agnostic to the spectral consistency observed within localized spatial regions in HSIs tend to perform poorly due to the heterogeneous spectral signatures and noise across an HSI scene~\cite{yang2021fuzzy, wang2017fast, chen2023spectral}. Spatially-regularized graphs counteract this spectral variance by limiting connections between pixels to those within a spatial radius around each pixel. Mathematically, a spatially-regularized kNN graph may be defined through its corresponding weight matrix $\mathbf{W}$, with $\mathbf{W}_{ij} = 1$ if $x_j$ is one of the $k_n$ Euclidean distance nearest neighbors of $x_i$ from points within a $(2R+1)\times (2R+1)$ spatial square centered at $x_i$ in the HSI, where $R\in \mathbb{N}$ is a user-defined spatial radius, and $\mathbf{W}_{ij} = 0$ otherwise~\cite{polk2023diffusion, murphy2019spectral, yang2021fuzzy, polk2021multiscale}. Thus, spatially-regularized graphs efficiently encode spatial coherence within the HSI by restricting edges within the graph underlying $\mathbf{W}$ to spatially-close pixels~\cite{zhai2021hyperspectral, polk2023diffusion, murphy2019spectral, yang2021fuzzy, polk2021multiscale}. 

Spatially-regularized graphs have emerged as a pivotal tool in enhancing HSI clustering and semi-supervised classification tasks~\cite{murphy2019spectral, polk2021multiscale, murphy2020spatially}. The spatially-regularized graph, specifically tailored for HSIs with diffusion distances and enhanced by a spatial window-based labeling consensus mechanism, was introduced in~\cite{murphy2019spectral} to effectively leverage spatial context. Extending this approach, a multiscale framework was introduced, utilizing spatially-regularized graphs to learn a single clustering scale that is most explanatory of latent multiscale cluster structure extracted by varying $t$ in diffusion distances, as measured by variation of information~\cite{polk2021multiscale, murphy2022multiscale}. Furthermore, spatially-regularized graphs have found utility in active learning, a branch of semi-supervised learning that requires human input. Spatially-regularized graphs aid in the strategic selection of a set of pixels for labeling---based on available budget---ensuring that the chosen pixels are locally coherent due to spatial constraints and globally representative through density estimation~\cite{murphy2020spatially}.

A second important field of research meant to incorporate spatial information into HSI clustering algorithms is that of superpixel segmentation. Superpixel segmentation algorithms partition the HSI into relatively small spatial regions exhibiting comparable spectral signatures, possibly due to similar mixtures of materials in the region they correspond to. These regions (or superpixels) capture local spatial structure, and the analysis of them (rather than that of the full HSI) reduces the number of pixels being analyzed, and hence, computational complexity associated with later analysis~\cite{liu2011entropy, wang2017superpixel}. Superpixel algorithms fall into two main classes: graph-based, which form superpixels by optimizing a cost function on a pixel-node graph~\cite{wang2017superpixel, FH2004FH, tang2012tpr, liu2011entropy, liu2017intrinsic}, and clustering-based, which iteratively cluster pixels based on convergence criteria~\cite{achanta2012SLIC, vincent1991watersheds, vedaldi2009quickshift, comaniciu2002meanshift, comaniciu2002meanshift}. Clustering-based superpixel segmentation algorithms may be used to produce superpixel segmentation at low computational cost~\cite{achanta2012SLIC}, but graph-based algorithms (such as ERS) often are capable of learning superpixel segmentations that better relate to latent manifold structure in an image~\cite{liu2011entropy}. 

Superpixel-based methods have been employed in HSI classification to incorporate spatial information into their predictions and reduce the effect of spatial-spectral variability within each superpixel~\cite{fang2015spectral, fang2015classification, sellars2020superpixel}. 
Indeed, ERS has been utilized to diminish the impact of noise on HSI clustering algorithms and to identify anchor points in HSI clustering~\cite{chen2023spectral}. Similarly, superpixel pooling autoencoders have been developed to capture superpixel-level latent representations to assist subsequent cluster analysis~\cite{cai2022superpixel}. Despite these developments, the integration and exploration of superpixels in unsupervised HSI clustering are still relatively limited, particularly in conjunction with diffusion geometry~\cite{bandyopadhyay2022tree, superBF, murphy2019spectral, polk2021multiscale, cui2022unsupervised}. 

\section{Superpixel-based and Spatially-regularized Diffusion Learning}
\label{sec:method}

\begin{figure}[ht]
    \centering
    \includegraphics[width = \textwidth]{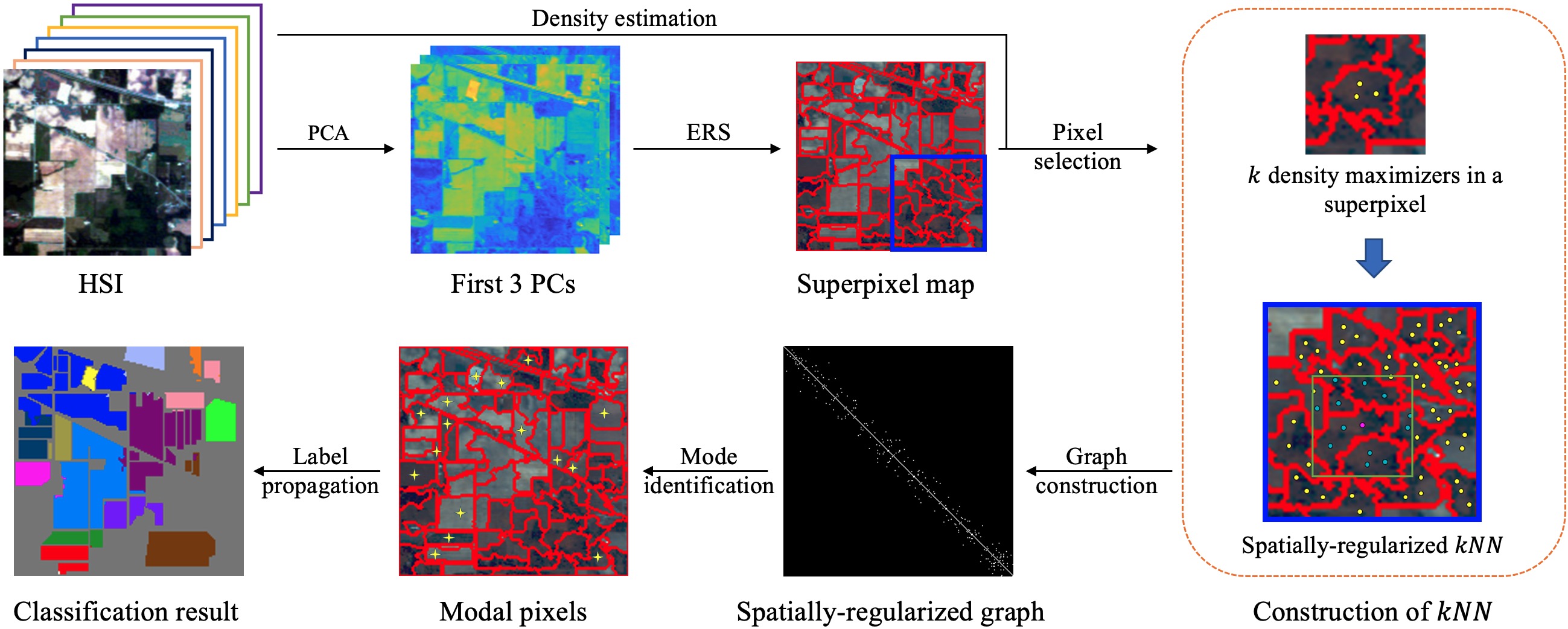}
    \caption{Workflow of the S$^2$DL Algorithm. The algorithm begins by projecting the first three PCs of the input HSI, which are then used to create a superpixel map via the ERS algorithm. S$^2$DL estimates the $k$ highest-density pixels within each superpixel as representatives in graph construction, and then constructs a spatially-regularized kNN graph. Mode pixels are subsequently identified and assigned unique labels, with the LBB of each mode receiving the same label as its respective mode. Then the labels are propagated to unlabeled selected pixels. The process concludes with majority voting within each superpixel to finalize the clustering.
    } 
    \label{fig:pipeline}
\end{figure}

This section introduces the proposed algorithm for unsupervised HSI clustering: Superpixel-based and Spatially-regularized Diffusion Learning (S$^2$DL). While pixel-wise clustering methods can be effective in certain scenarios, they often fail when applied to HSIs due to their disregard for spatial information: a key characteristic in HSI data. Ignoring the spatial context can lead to inaccurate clustering results as the spatial proximity of pixels often correlates with similarity in material composition in remotely-sensed images. S$^2$DL addresses this challenge by integrating spatial information into its clustering process in two key steps. First, S$^2$DL employs the graph-based ERS superpixel segmentation technique to obtain a high-quality superpixel segmentation of the HSI~\cite{liu2011entropy, xie2019SLICshortage}, as described in Section \ref{sec: ERS}. Second, as described in Section \ref{sec: graph}, S$^2$DL constructs a spatially-regularized graph from carefully-selected pixels from each ERS-derived superpixel, effectively capturing intrinsic spectral-spatial relationships between pixels within the HSI~\cite{murphy2019spectral, murphy2018unsupervised, polk2021multiscale} while reducing overall computational complexity of graph construction and analysis. Finally, S$^2$DL performs a diffusion-based clustering procedure to label superpixel exemplars and propagate those labels to the remaining image as described in Section \ref{sec: Diffusion Learning}. The combination of superpixel segmentation and the use of a spatially-regularized graph is expected to result in superior clustering results using S$^2$DL (as will be shown in our numerical experiments in Section \ref{sec: results}). As is shown in the complexity analysis in Section \ref{sec: complexity}, S$^2$DL's use of superpixel segmentation and spatially-regularized diffusion distances offers a computationally efficient approach to HSI cluster analysis that is expected to translate to high-quality unsupervised analysis of large-scale HSI datasets. The detailed steps of S$^2$DL are introduced in the following sections and summarized in Figure \ref{fig:pipeline} and Algorithm \ref{alg:s2dl}.

\subsection{ERS-Based Superpixel Segmentation}\label{sec: ERS}

In its first step, the proposed S$^2$DL computes a superpixel segmentation map using ERS; see Algorithm~\ref{ERS}~\cite{liu2011entropy}. As will become clear soon, the superpixels derived using ERS not only reflect spectral similarities between pixels in a graph-based procedure, but also the spatial information. Incorporating spatial information through superpixels offers a significant improvement over traditional pixel-wise clustering techniques that often disregard spatial information.

ERS builds its superpixel segmentation based on an undirected weighted graph based on the projection of pixel spectra onto their first three principal components (PCs), denoted $X_{PCA}$. Mathematically, this initial graph may be defined using the weight matrix $\mathbf{Z}\in\mathbb{R}^{N\times N}$ with  
\[
\mathbf{Z}_{ij} = \begin{cases} 
    \exp\left(-\frac{\|l\left(x_i\right)-l\left(x_j\right)\|_2^2\cdot \|x_i-x_j\|_2^2}{2\sigma^2}\right) & \text{if } x_i\in \mathcal{N}_\ell(x_i) \text{ or } x_j\in \mathcal{N}_\ell(x_j), \\
    0 & \text{otherwise,}
\end{cases}
\]
where $l\left(x_i\right)$ and $l\left(x_j\right)$ are spatial coordinates of pixels $x_i$ and $x_j$, respectively, and $\sigma>0$ is a tuning parameter meant to control interactions between pixels, and $N_\ell(x)$ is the set of $\ell$ spatial nearest neighbors of $x$ in the PCA-reduced feature space. In our later experiments, we set $\sigma = 5$ and $\ell=8$: the default values for these parameters~\cite{liu2011entropy}.  Denote the edge set underlying this graph as $E = \{(i,j) | \mathbf{Z}_{ij}>0\}$. ERS performs superpixel segmentation by locating a subset of edges $A \subseteq E$ to form $N_s$ compact, homogeneous, and well-balanced superpixels using the following optimization:
\begin{equation} \label{eq: ERS}
A^* = \argmax\limits_{A} \mathcal{J}(A) = \mathcal{H}(A) + \alpha \mathcal{B}(A), \quad \text{s.t. } A \subseteq E \text{ and } N_A \geq N_s,
\end{equation}
where $N_A$ is the number of connected components in $A$ and $\alpha >0$ is the balancing factor that determines the trade-off between the two terms $\mathcal{H}(A)$ and $\mathcal{B}(A)$, which are described below. 

First, the term $\mathcal{H}(A)=-\sum_{i=1}^{N} \bm{\mu}_{i} \sum_{j=1}^N q_{ij}(A) \log \left(q_{ij}(A)\right)$ measures the entropy rate of the edge set $A$, encouraging compact and homogeneous spatial regions in the ERS superpixel segmentation by considering a random walk on the graph subset $A$~\cite{liu2011entropy}. The entropy rate calculation relies on two main quantities: the $N\times 1$ stationary distribution $\bm{\mu}\in\mathbb{R}^{N\times 1}$ of a random walk on the initial graph underlying $\mathbf{Z}$---defined by $\bm{\mu}_i = \frac{\sum_{i=1}^N \mathbf{Z}_{ij}}{\sum_{i,j=1}^N \mathbf{Z}_{ij}}$---and transition probabilities $q_{ij}(A)$ on the edge subset $A$ of $E$, defined as 
\[
q_{ij}(A) = \begin{cases}
    1 - \frac{\sum_{k\in A_i} \mathbf{Z}_{ik}}{\sum_{k=1}^N \mathbf{Z}_{ik}} & \text{if } i = j,\\
    \frac{\mathbf{Z}_{ij}}{\sum_{k=1}^N \mathbf{Z}_{ik}} & \text{if } i \neq j \text{ and } (i,j)\in A, \\
    0 & \text{if } i \neq j \text{ and } (i,j) \notin A,
\end{cases}
\]
where $A_i = \{k\in \{1,2,\dots, N\}|(i,k)\in A\}$ denotes the set of indices of pixels directly connected to $x_i$ in $A$. 
Thus, the entropy rate $\mathcal{H}(A)$ increases monotonically with the addition of edges to $A$, particularly when these edges contribute to more coherent clusters~\cite{liu2011entropy}. 

The second term in ERS's optimization $\mathcal{B}(A)$ is meant to encourage uniform superpixel size and a balanced number of superpixels. Mathematically,  $\mathcal{B}(A)=-\sum_{i=1}^{N_A} \bm{r}_i\log\left(\bm{r}_i\right)-N_A$, where $\mathbf{r}\in[0,1]^{N_A}$ denotes the distribution of pixels in $N_A$ connected components; i.e., $\bm{r}_i$ is the fraction of pixels in $X_{PCA}$ that exist in connected component $i$. Therefore, all else equal, the incorporation of $\mathcal{B}(A)$ downweights both complicated superpixel segmentations with high variation in superpixel size across the image, or segmentations with extraneous superpixels. Hence, the incorporation of $\mathcal{B}(A)$ can be interpreted as balancing the resulting ERS superpixel segmentation, ensuring that new superpixels are roughly equal in scale when edges are added to $A$ in the ERS optimization~\cite{liu2011entropy}.

Together, these two terms in $\mathcal{J}(A)$ monotonically increase as edges are added to $A$, referencing uniformly-sized, spectrally-similar spatial regions as derived superpixels. The proposed objective function is efficiently addressed using a ``lazy greedy'' heuristic that initializes with an empty edge set $A$ and gradually adds edges to maximize the ERS objective function~\cite{nemhauser1978greedy}. This iterative addition persists until the number of connected components $N_A$ matches the pre-set target $N_s$, thereby achieving the specified number of superpixels~\cite{liu2011entropy}. Finally, superpixels are generated from the connected edges in the optimized edge set $A^*$. Mathematically, a superpixel $S_i$ is a set of pixels with a coherent subset of edges in $A^*$; see details in Algorithm~\ref{ERS}. In the proposed S$^2$DL algorithm, the superpixel segmentation $\mathcal{S}$ is utilized for subsequent graph construction, as will become clear soon in Section \ref{sec: graph}. 

\begin{algorithm}
\renewcommand{\baselinestretch}{1.5}\selectfont
\DontPrintSemicolon

\let\oldnl\nl
\newcommand{\nonl}{\renewcommand{\nl}{\let\nl\oldnl}}
\SetAlgoLined
\Input{$X_{PCA}$ (Projection of pixels onto first 3 PCs), $N_s$ ($\#$ superpixels), $\alpha$ ($\#$ balancing factor)}
\Output{$S$ (Superpixel Map)}

Construct the graph $G=(X_{PCA},E,\mathbf{Z})$\;
Initialize the subset edge $A^*=\emptyset$, $U=E$ and $S=\{S_i\mid S_i=\{x_i\}, 1\leq i\leq N\}$;

\While{$U\neq \emptyset$ \rm{\textbf{and}} $N_A\geq N_s$} {
Find edge $\hat{a} = \argmax_a\mathcal{J}(A^*\cup \{a\})-\mathcal{J}(A^*)$, where $\mathcal{J}(A^*)=\mathcal{H}(A^*)+\alpha \mathcal{B}(A^*)$

\If{$A^*\cup\{\hat{a}\}$ \rm{is cycle-free} }{
Update $A^*=A^*\cup\{\hat{a}\}$}

Update $U=U-\{\hat{a}\}$

}
\While{$A\neq \emptyset$} {
\If{$(i,j)\in A$}{
Find $x_i\in S_a$ and $x_j \in S_b$, update $S_a = S_a\cup S_b$ \\
Update $A=A-\{(i,j)\}$ and $S=S-{S_b}$}

}
\caption{Entropy Rate Superpixel (ERS) algorithm~\cite{liu2011entropy}}
\label{ERS}
\end{algorithm}

\subsection{Reduced Spatially Regularized Graph Construction} \label{sec: graph}

S$^2$DL computes a spatially regularized graph (see Section \ref{sec: spectral-spatial} for details)  using a small subset of carefully-selected pixels from the ERS superpixel segmentation. To find the representative pixels from each superpixel, S$^2$DL first relies on the following quantity: 
\begin{equation}
    \zeta(x) = \frac{1}{\mathcal{Z}}\sum_{y\in {k_n}(x)}\exp(-||x-y||_2^2/\sigma_0^2)
\end{equation}
for each pixel $x\in X$, where ${k_n}(x)$ denotes the set of $k_n$ $\ell^2$-nearest neighbors of the HSI pixel $x$ in $X$, $\sigma_0>0$ is a scaling factor controlling the interaction radius between pixels, and the quantity $\mathcal{Z}$ normalizes $\zeta(x)$ to ensure that $\sum_{y \in X} \zeta(y) = 1$. Thus, $\zeta(x)$ will be higher for modal pixels that are close in Euclidean distance to their $k_n$ $\ell^2$-nearest neighbors and small otherwise~\cite{polk2023diffusion, murphy2018unsupervised, maggioni2019learning}. For each ERS superpixel, S$^2$DL selects the $k$ pixels within the superpixel maximizing $\zeta(x)$, resulting in $N_s \times k$ representative pixels used for graph construction. Mathematically, we define this highly-explanatory subset $X_s$ of $X$ by $X_s = \bigcup_{i=1}^{N_s}\{x\in S_i | \text{$x$ is one of the $k$ maximizers of $\zeta(x)$ in $S_i$}\}$. Significantly,  using $X_s$ rather than $X$ in graph construction results in a substantial reduction in computational complexity, while maintaining an accurate characterization of essential spatial-spectral and geometric information. Once representative pixels have been identified, a sparse spatially-regularized kNN adjacency graph is constructed from the pixels in $X_s$, following the procedure described in Section \ref{sec: spectral-spatial}. This graph encodes rich spatial information in an HSI data by restricting edges between pixels to spatial nearest neighbors~\cite{polk2021multiscale, murphy2020spatially}. 

\subsection{Diffusion-Based Clustering}\label{sec: Diffusion Learning}

This section describes S$^2$DL's diffusion-based clustering procedure to obtain high-quality labeling of superpixels. S$^2$DL is inspired by the ubiquitous DPC algorithm~\cite{rodriguez2014clustering} (See Section \ref{sec:clustering}) and its diffusion geometry-based extensions~\cite{murphy2018unsupervised, maggioni2019learning, polk2023diffusion} that rely on diffusion distances rather than Euclidean distances to perform cluster analysis. Specifically, S$^2$DL uses diffusion distances calculated from the spatially-regularized graph introduced in Section \ref{sec: graph} for mode selection~\cite{maggioni2019learning, murphy2019spectral, polk2021multiscale}. First, S$^2$DL locates $K$ pixels meant to serve as cluster modes---exemplars for latent cluster structure---and assigns these pixels unique labels. These cluster modes are identified as the $K$ pixels in $X_s$ that maximize $\Delta_t(x) = d_t(x)\zeta(x)$, where 
\begin{equation} \label{eq: dt}
   d_t(x) = 
   \begin{cases} 
       \max_{y \in X_s} D_t(x,y) & x = \argmax_{y\in X_s} \zeta(y), \\ 
       \min_{y\in X_s}\{ D_t(x,y) | \zeta(y)\geq \zeta(x)\} & \text{otherwise}.
   \end{cases}
\end{equation}
In particular, $d_t(x)$ is the diffusion distance at time $t$ between $x$ and that HSI pixel's $D_t$-nearest neighbor with higher density. Thus, cluster modes identified using S$^2$DL are the highest-density pixels in $X_s$ farthest in diffusion distance from other highest-density pixels~\cite{murphy2018unsupervised, maggioni2019learning, polk2021multiscale, polk2022unsupervised, polk2023diffusion, murphy2019spectral, murphy2020spatially, murphy2022multiscale}. S$^2$DL assigns $\hat{C}(x_{m_k}) = k$ for $k=1,2, \dots, K$, where $\{x_{m_k}\}_{k=1}^K$ are the $K$ maximizers of $\Delta_t(x)$ and $\hat{C}\in\{1,2,\dots,K\}^N$ is a clustering map with $\hat{C}_i= k$ indicating that pixel $x_i$ is assigned to cluster $k$.

After locating cluster modes, S$^2$DL relies on a local backbone (LBB) to propagate modal labels to unlabeled pixels in $X_s$~\cite{seyedi2019dynamic}. An LBB generally consists of a cluster center and points that have the highest probability of sharing the same label with that cluster center, as determined by similarity measures such as Euclidean distance or cosine similarity. In S$^2$DL, the LBB of each modal pixel---defined as the first ${k_n}$ spatially-regularized nearest neighbors of that cluster mode---is assigned the same label as that of the modal pixel~\cite{seyedi2019dynamic}. The LBB's formation is of significance in the S$^2$DL algorithm as it emphasizes the spatial coherence of the HSI data at the beginning of its non-modal labeling procedure. Next, in order of descending density, the remaining pixels $x\in X_s$ are labeled according to their $D_t$-nearest neighbor of higher density that is already labeled~\cite{maggioni2019learning, rodriguez2014clustering, murphy2019spectral}: $\hat{C}(x) = \hat{C}({x}^*)$, where 
\begin{equation} \label{eq: nonmodal labeling}
    {x}^* = \argmin\limits_{y\in X_s}\{D_t(x,y)|
    \hat{C}(y)>0
     \ \land \ \zeta(y)\geq \zeta(x) \}.
\end{equation}
Once all pixels in $X_s$ have been labeled, S$^2$DL propagates labels within to the rest of the image using a majority voting process: assigning the majority label of representative pixels in each superpixel to all pixels in a superpixel. Notably, this majority voting procedure further enforces the retention of essential spatial-spectral characteristics of the original HSI data in S$^2$DL.

\begin{algorithm}
\renewcommand{\baselinestretch}{1.5}\selectfont
\DontPrintSemicolon

\let\oldnl\nl
\newcommand{\nonl}{\renewcommand{\nl}{\let\nl\oldnl}}

\Input{$X$ (HSI), $N_s$ ($\#$ superpixels), $k$ ($\#$ representative pixels per superpixel), $\sigma_0$ (kernel scaling factor), $k_n$ ($\#$ nearest neighbors), $R$ (spatial radius), $K$ ($\#$ of clusters)}
\Output{$C$ (Clustering Map)}

\nonl\textbf{ERS-Based Superpixel Segmentation:}\;
\Indp 
Calculate $X_s$: the projection of pixels in $X$ onto its first three PCs\;
Run ERS to segment the PCA-reduced HSI $X_{PCA}$ into \(N_s\) superpixels\;
\Indm

\nonl\textbf{Reduced Spatially Regularized Graph Construction:}\;
\Indp 
Compute kernel density estimation \(\zeta(x) = \sum_{y\in {k_n}(x)}\exp(-||x-y||_2^2/\sigma_0^2)\) for all $x\in X$\;
For each superpixel, store the $k$ pixels in that superpixel maximizing $\zeta(x)$ in $X_s$\; 
Construct a spatially-regularized kNN adjacency graph using the selected $k\cdot N_s$ pixels\ with spatial radius $R$\;
\Indm

\nonl\textbf{Diffusion-Based Clustering:}\;
\Indp 
Compute \(\Delta_t(x) = \zeta(x)d_t(x)\) for $x\in X_s$, where  \(d_t(x)\) is as in Equation \ref{eq: dt}\;  
Identify the \(K\) maximizers of \(\Delta_t(x)\) as modal pixels and assign unique labels from 1 to $K$\;
For each cluster mode, assign its local backbone (with $k_n$ nearest neighbors) the modal pixel's label\;  
In order of decreasing density, assign each unlabeled pixel in $X_s$ the label of their \(D_t\)-nearest neighbor of higher density \(\zeta\) that is already labeled (Equation \ref{eq: nonmodal labeling})\;
For each superpixel, assign all pixels in the superpixel the modal label among the $k$ representative pixels in $X_s$ from this superpixel\;
\Indm

\caption{Superpixel-based and Spatially-regularized Diffusion Learning (S$^2$DL) method}
\label{alg:s2dl}
\end{algorithm}

\subsection{Computational Complexity}\label{sec: complexity}

This section analyzes S$^2$DL (Algorithm~\ref{alg:s2dl}) concerning its computational complexity and scaling. The first stage, wherein superpixels are calculated using ERS scales has the following main components: $\mathcal{O}(NB^2)$ to calculate the first three principal components. While ERS's worst-case complexity is $\mathcal{O}(N^2\log(N))$, practical implementations typically exhibit faster performance, often leading to an average-case complexity of $\mathcal{O}(N\log(N))$~\cite{liu2011entropy}.

For nearest neighbor searches, we assume the use of cover trees, which enables efficient nearest neighbor searches in high-dimensional spaces~\cite{beygelzimer2006cover}. Indeed, the computational complexity of searching for the $k$ $\ell^2$-nearest neighbors in $X$ using cover trees is $\mathcal{O}(k_n Bc^{d} N \log(N))$, where $d$ is the HSI's doubling dimension~\cite{beygelzimer2006cover, polk2023diffusion} and $c>0$ is a constant that is $\mathcal{O}(1)$ with respect to the other parameters $k_n$, $B$, $d$, and $N$. In this complexity analysis, we assume that these two values remain constant across subsets of $X$; e.g., that the doubling dimension of $X$ is the same as that of $X_s$~\cite{beygelzimer2006cover}. Under this assumption, the computational complexity of computing the KDE at each pixel is $\mathcal{O}(k_n Bc^{d} N \log(N))$. Similarly, locating the $k$ KDE-maximizers from each superpixel has complexity $\mathcal{O}(N_s\log(N/N_s) + \mathcal{O}(kN_s)$. Finally, the computational complexity of building our reduced spatially regularized graph is $\mathcal{O}(k_n Bc^{d} kN_s \log(kN_s))$: a notable reduction in computational complexity due to our earlier downsampling procedure. 

To perform its diffusion-based clustering, S$^2$DL requires $\mathcal{O}(k_n B c^d k N_s \log(k N_s) + k_n L^2 k N_s )$ to calculate $d_t(x)$, where $L$ is the number of eigenvectors of $\mathbf{P}$ used to approximate diffusion distances~\cite{maggioni2019learning, coifman2006diffusion}. In contrast to diffusion-based methods clustering all the pixels, S$^2$DL therefore operates at a significantly reduced computational complexity through its analysis of superpixel exemplars. Then, the computational complexity of labeling the LBB and remaining pixels is $\mathcal{O}(Kk_n)$ and $\mathcal{O}(k_n LBc^{d} kN_s \log(kN_s))$, respectively~\cite{seyedi2019dynamic, maggioni2019learning}. Finally, it costs $\mathcal{O}(N_s k)$ to perform majority voting. Prior work has demonstrated that $k_n = \mathcal{O}(\log(N))$~\cite{murphy2022multiscale, maggioni2019learning, polk2023diffusion} and we expect that, for HSIs taken over the same scene, $N_s = \mathcal{O}(1)$ and $K= \mathcal{O}(1)$ with respect to $N$. Assuming that $L = \mathcal{O}(1)$ and $k = \mathcal{O}(\log(N))$ with respect to $N$ also, S$^2$DL's overall computational complexity reduces to $\mathcal{O}(Bc^{d} N\log^2(N))$: log-linear with respect to the number of pixels. Notably, S$^2$DL's computational complexity is dominated by its calculation of the KDE (not graph construction, as in other diffusion-based algorithms), indicating high-quality scaling to large-scale HSI clustering problems.

\section{Experimental Results and Discussion}\label{sec: results}

This section contains extensive numerical experiments showing the efficacy of the proposed S$^2$DL algorithm. We compared S$^2$DL against both traditional and state-of-the-art comparison methods on three real-world HSIs often used for benchmarking new algorithms (Section \ref{sec:benchmark}). In addition, we demonstrate that  S$^2$DL may be applied to the real-world problem of landscape-scale species mapping of mangroves using remotely-sensed HSI data collected over the Mai Po Nature Reserve (Section \ref{sec:hk}).

Among the algorithms we analyzed, several serve as baselines, as they do not integrate spatial information. These include: $K$-Means~\cite{kmeans}, which partitions data by minimizing intra-cluster $\ell^2$-distances.  Spectral Clustering (SC)~\cite{ng2001spectral}, which employs $K$-Means on the first $K$ eigenvectors of $\mathbf{P}$. DPC defines cluster modes as high-density points distant from others, assigning them unique labels that are then propagated to neighboring points based on decreasing density~\cite{rodriguez2014clustering}. PGDPC~\cite{guan2021peak} discriminates many ``peak'' pixels within the image with highest density among nearest neighbors as measured with a KDE from and ``non-peak'' pixels, which are associated with their nearest neighbor of higher density. Graph-based cluster assignments are derived for peak pixels (relying on both density and pairwise geodesic distances for graph construction) and propagated along geodesic paths to remaining non-peak pixels~\cite{guan2021peak}.  Diffusion  Learning (DL)~\cite{maggioni2019learning}---recognized as an early diffusion-based clustering technique---relies on the metric $\Delta_t(x)$ from Section \ref{sec: Diffusion Learning} for cluster mode identification and label propagation.  Diffusion and Volume Maximization-based Image Clustering (D-VIC)~\cite{polk2023diffusion} is a recent diffusion-based clustering algorithm that incorporates spectral unmixing into a DL clustering framework, downweighting high-density, low-purity pixels in mode selection and non-modal labeling. Although shown to be successful on a wide range of  HSI data~\cite{murphy2018unsupervised, murphy2022multiscale, polk2023diffusion}, DL and D-VIC do not natively incorporate spatial information into their labeling procedures.

Other comparison algorithms implemented exploit both spatial and spectral information in their labeling procedures. Improved Spectral Clustering with Multiplicative Update Algorithm (SC-I)~\cite{zhao2019fast} modifies SC by iteratively solving the eigenvalue decomposition of the Laplacian matrix $\mathbf{L}$, relaxing its discreteness condition, and integrating spatial context into the graph underlying $\mathbf{P}$. SLIC-PGDPC (S-PGDPC)~\cite{guan2021peak} extends PGDPC by taking the average of the spectral signatures within each superpixel and using this average as input for the PGDPC algorithm. Spectral-Spatial Diffusion Learning (DLSS)~\cite{murphy2018unsupervised} enhances DL by adopting a two-stage labeling scheme: initially, it assigns labels via DL, subject to a spatial consensus check through majority voting within a specified window, leaving non-conforming pixels unlabeled; this is followed by a secondary DL process to label the remaining unlabeled pixels. Spatial-Spectral Image Reconstruction and Clustering with Diffusion Geometry (DSIRC) builds on the D-VIC framework, utilizing both purity and density metrics for cluster mode identification, and enhances clustering accuracy by integrating spatial information through a shape-adaptive reconstruction process that effectively reduces noise before applying Diffusion Learning~\cite{cui2022unsupervised}. Spatially-Regularized Diffusion Learning (SRDL)~\cite{murphy2019spectral, polk2021multiscale} further refines DLSS by constructing a spatially-regularized graph, resulting in clusters that are more spatially consistent. Hyperparameters for all algorithms were optimized for across a grid search, as is described in Appendix \ref{app: hyperparameter}.

To evaluate the performance of our S$^2$DL algorithm, we employ a suite of metrics.  Overall Accuracy (OA) calculates the total fraction of pixels correctly clustered. Average Accuracy (AA) measures the average OA across different classes. Cohen's kappa coefficient ($\kappa$), defined as $\kappa = \frac{p_{o}-p_{e}}{1-p_{e}}$, contrasts observed accuracy against expected random accuracy~\cite{cohen1960coefficient}. Additionally, we track the runtime (RT) in seconds to assess computational efficiency. For all datasets, we set $K$ as the ground truth number of clusters. All experiments were conducted in MATLAB R2021a with the same environment: Intel® Core™ i7-10875H CPU @ 2.30GHz, 8 cores, 64 GB RAM, run on a Windows 64-bit system. The optimal output was defined as the clustering that maximizes the sum of OA, AA and $\kappa$. The code to replicate numerical experiments can be found at: \href{https://github.com/ckn3/S2DL}{github.com/ckn3/S2DL}.

\subsection{Experiments on Benchmark HSI Datasets}

\label{sec:benchmark}

\subsubsection{Benchmark Datasets}

In this section, we introduce the three benchmark datasets chosen for this study. These datasets, captured using the AVIRIS sensor, serve as representatives of diverse agricultural landscapes and have been widely utilized for evaluating machine learning methods for HSI.

\noindent \textbf{Salinas and Salinas A}: Captured by the AVIRIS sensor in 1998, the spatially-regular dataset Salinas showcases the agricultural terrains of Salinas Valley, California. It has a spectral range of 380~nm to 2500~nm across 224 bands, with spatial size $512\times 217$ pixels, totaling 111104 pixels. The Salinas A subset zooms in on a specific region of the Salinas scene with $83\times 86$ pixels, totaling 7138 pixels. Gaussian noise was added for distinctiveness. While the broader Salinas dataset has 16 classes, Salinas A focuses on 6 main crop types.

\noindent \textbf{Indian Pines}: Produced by the AVIRIS sensor in 1992, this dataset portrays northwest Indiana farmlands. It covers a spectral range of 400~nm to 2500~nm over 224 bands and spreads over $145\times 145$ pixels, making up 21025 pixels. Notably, it includes 16 ground truth classes, capturing various crops and infrastructure.

\subsubsection{Numerical Results on Benchmark Datasets}

This section offers detailed comparisons between the clusterings produced by S$^2$DL and various classical and state-of-the-art algorithms introduced earlier in this section. Across the three benchmark HSIs analyzed, S$^2$DL delivers clusterings with the highest performance in terms of three evaluation metrics. Notably, on the Salinas A dataset, the method achieves near-perfect clustering performance, despite its unsupervised setting. Similarly, although the Indian Pines dataset is widely considered challenging due to its many classes being distributed widely across the scene, S$^2$DL still manages to surpass its competitors in performance. Finally, on the Salinas dataset, S$^2$DL not only yields the best performance in OA and $\kappa$ but also completes the task in 8.80 seconds of runtime. In contrast, the next best algorithm (SRDL) required 445.31 seconds of runtime. This impressive improvement on runtime is attributed to the use of superpixel-based reduction in graph size, illustrating S$^2$DL's balanced approach to both speed and accuracy. These findings underscore S$^2$DL's efficacy and efficiency in HSI clustering, marking it as an suitable choice for practical applications.

\begin{table}[ht]
\centering
\caption{Comparison of Unsupervised Clustering Methods on Benchmark HSI Datasets. The best and second-best performances are indicated by bold and underlined values, respectively. S$^2$DL stands out for its high-quality clustering across almost all metrics and datasets, except for the AA value on Salinas. Its relatively low runtimes compared to other algorithms also highlight its scalability for analysis of large HSIs.} 
\vspace{0.2cm}

\label{tab:datasets}
\resizebox{\textwidth}{!}{%
\begin{tabular}{|l|ccccccccccccc|}
\hline
\multirow{2}{*}{\textbf{Dataset}} & \multicolumn{13}{c|}{\textbf{Method}} \\ \cline{2-14}
 &          & $K$-Means & SC & DPC & PGDPC & DL & D-VIC & SC-I  &S-PGDPC& DLSS  & DSIRC & SRDL & S$^2$DL \\ \hline
\multirow{4}{*}{Salinas A} 
 & OA       & 0.764 & 0.841 & 0.786 & 0.844 & 0.887 & \underline{0.976} & 0.827 & 0.647 & 0.890 & 0.911 & 0.895 & \textbf{0.996} \\ 
 & AA       & 0.749 & 0.887 & 0.849 & 0.893 & 0.920 & \underline{0.973} & 0.875 & 0.680 & 0.888 & 0.903 & 0.926 & \textbf{0.996} \\ 
 & $\kappa$ & 0.703 & 0.806 & 0.740 & 0.813 & 0.860 & \underline{0.970} & 0.789 & 0.568 & 0.862 & 0.889 & 0.870 & \textbf{0.995} \\
 & RT       & 0.05  & 1.59  & 2.66  & 1.63  & 1.93  & 4.89              & 6.43  & 0.10  & 5.27  & 26.39 & 14.99 & 1.78  \\ \hline
\multirow{4}{*}{Indian Pines} 
 & OA       & 0.386 & 0.382 & 0.391 & 0.428 & 0.404 & 0.471 & 0.496 & 0.477 & 0.467 & 0.620 & \underline{0.640} & \textbf{0.647} \\ 
 & AA       & 0.398 & 0.368 & 0.376 & 0.399 & 0.401 & 0.376 & 0.304 & 0.530 & 0.462 & 0.549 & \underline{0.553} & \textbf{0.591} \\ 
 & $\kappa$ & 0.315 & 0.313 & 0.304 & 0.351 & 0.313 & 0.383 & 0.394 & 0.431 & 0.400 & 0.573 & \underline{0.596} & \textbf{0.602} \\
 & RT       & 1.14  & 14.40 & 13.10 & 19.75 & 13.64 & 24.33 & 70.43 & 0.38  & 20.55 & 136.02& 30.52 & 2.19  \\ \hline
\multirow{4}{*}{Salinas}  
 & OA       & 0.639 & 0.662 & 0.668 & --    & 0.687 & 0.696 & --    & 0.590 & 0.702 & 0.677 & \underline{0.834} & \textbf{0.887} \\ 
 & AA       & 0.612 & 0.633 & 0.654 & --    & 0.662 & 0.623 & --    & 0.487 & 0.674 & 0.612 & \textbf{0.756} & \underline{0.729} \\ 
 & $\kappa$ & 0.597 & 0.620 & 0.627 & --    & 0.646 & 0.653 & --    & 0.551 & 0.662 & 0.633 & \underline{0.813} & \textbf{0.874} \\
 & RT       & 4.81  & 414.44& 432.98& --    & 450.82& 496.37& --    & 1.20  & 504.88&3059.74& 445.31& 8.80  \\ \hline
\end{tabular}%
}
\end{table}

\begin{figure*}[ht]
    \centering
    
    \begin{subfigure}[t]{0.135\textwidth}
    \centering
    \includegraphics[width = \textwidth]{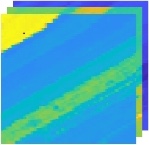} \hspace{0.02in}
    \vspace{-0.5cm}
    \caption{First 3 PCs}
    \end{subfigure}
    \begin{subfigure}[t]{0.135\textwidth}
    \centering
    \includegraphics[width = \textwidth]{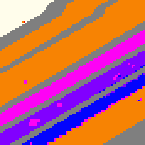} \hspace{0.02in}
    \vspace{-0.5cm}
    \caption{$K$-Means}
    \end{subfigure}
    \begin{subfigure}[t]{0.135\textwidth}
    \centering
    \includegraphics[width = \textwidth]{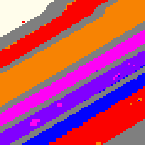} \hspace{0.02in}
    \vspace{-0.5cm}
    \caption{SC}
    \end{subfigure}
    \begin{subfigure}[t]{0.135\textwidth}
    \centering
    \includegraphics[width = \textwidth]{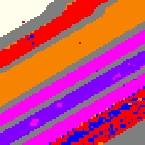} \hspace{0.02in}
    \vspace{-0.5cm}
    \caption{DPC}
    \end{subfigure}
    \begin{subfigure}[t]{0.135\textwidth}
    \centering
    \includegraphics[width = \textwidth]{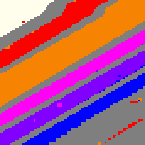} \hspace{0.02in}
    \vspace{-0.5cm}
    \caption{PGDPC}
    \end{subfigure}
    \begin{subfigure}[t]{0.135\textwidth}
    \centering
    \includegraphics[width = \textwidth]{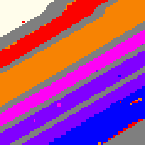} \hspace{0.02in}
    \vspace{-0.5cm}
    \caption{DL}
    \end{subfigure}
    \begin{subfigure}[t]{0.135\textwidth}
    \centering
    \includegraphics[width = \textwidth]{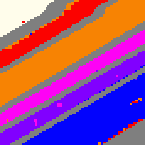} 
    \vspace{-0.5cm}
    \caption{D-VIC}
    \end{subfigure}
    
    \begin{subfigure}[t]{0.135\textwidth}
    \centering
    \includegraphics[width = \textwidth]{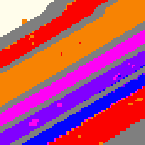} \hspace{0.02in}
    \vspace{-0.5cm}
    \caption{SC-I}
    \end{subfigure}
    \begin{subfigure}[t]{0.135\textwidth}
    \centering
    \includegraphics[width = \textwidth]{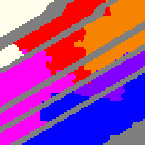} \hspace{0.02in}
    \vspace{-0.5cm}
    \caption{S-PGDPC}
    \end{subfigure}
    \begin{subfigure}[t]{0.135\textwidth}
    \centering
    \includegraphics[width = \textwidth]{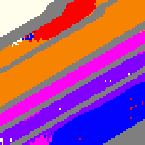} \hspace{0.02in}
    \vspace{-0.5cm}
    \caption{DLSS}
    \end{subfigure}
    \begin{subfigure}[t]{0.135\textwidth}
    \centering
    \includegraphics[width = \textwidth]{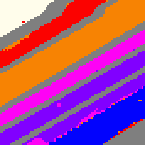} \hspace{0.02in}
    \vspace{-0.5cm}
    \caption{DSIRC}
    \end{subfigure}
    \begin{subfigure}[t]{0.135\textwidth}
    \centering
    \includegraphics[width = \textwidth]{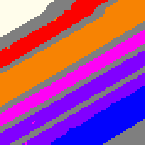} \hspace{0.02in}
    \vspace{-0.5cm}
    \caption{SRDL}
    \end{subfigure}
    \begin{subfigure}[t]{0.135\textwidth}
    \centering
    \includegraphics[width = \textwidth]{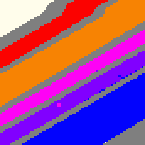} \hspace{0.02in}
    \vspace{-0.5cm}
    \caption{S$^2$DL}
    \end{subfigure}
    \begin{subfigure}[t]{0.135\textwidth}
    \centering
    \includegraphics[width = \textwidth]{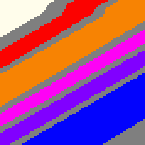} 
    \vspace{-0.5cm}
    \caption{Ground Truth}
    \end{subfigure}
    
    \caption{Comparison of clustering results of algorithms using spectral information (panels (b)-(g)), and algorithms using both spatial and spectral information (panels (h)-(l)) and S$^2$DL (panel (m)) on the Salinas A dataset (panel (a) with ground truth (panel(n))).}
    \label{fig:results_saa}
\end{figure*}

As visualized in Figure \ref{fig:results_saa}, S$^2$DL demonstrates notable precision in accurately identifying the ground truth labels within the Salinas A dataset. Whereas various algorithms split the cluster associated with the 8-week maturity romaine into two parts (visualized in dark blue in Figure \ref{fig:results_saa}), both S$^2$DL and the next highest-performing algorithm---D-VIC---correctly group these pixels into a unified cluster, with S$^2$DL surpassing D-VIC by approximately 2\% across all three metrics (see Table \ref{tab:datasets}). The precision demonstrated by D-VIC primarily stems from its reliance on spectral unmixing information~\cite{polk2023diffusion}, whereas S$^2$DL’s effectiveness is attributed to its incorporation of both spatial and spectral data into its diffusion-based clustering procedure. Notably, S$^2$DL mitigates spatial noise in its clustering through its superpixelation step, yielding a more spatially-regularized clustering compared to D-VIC. This demonstrates the capability of S$^2$DL to utilize spatial information as a robust alternative to the spectral unmixing in D-VIC, especially with spatially-regular HSIs.

The performance of S$^2$DL on the Salinas dataset is particularly noteworthy, having achieved highest OA and $\kappa$ scores in Table \ref{tab:datasets}.  The AA of S$^2$DL is marginally lower than that of SRDL (<3\%), potentially because of SRDL's efficiency in recognizing smaller classes. Nevertheless, this slightly lower AA is compensated with over 5\% higher OA and $\kappa$. Additionally, S$^2$DL boasts a much lower runtime due to its superpixelation step, further highlighting its efficiency. The gap in runtime between S$^2$DL, SRDL, and other methods indicates the significance of a spatially-regularized graph approach in handling large and regular HSIs like Salinas. 

\begin{figure*}[ht]
    \centering
    
    \begin{subfigure}[t]{0.135\textwidth}
    \centering
    \includegraphics[width = \textwidth]{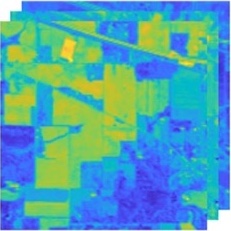} \hspace{0.02in}
    \vspace{-0.5cm}
    \caption{First 3 PCs}
    \end{subfigure}
    \begin{subfigure}[t]{0.135\textwidth}
    \centering
    \includegraphics[width = \textwidth]{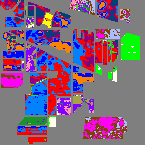} \hspace{0.02in}
    \vspace{-0.5cm}
    \caption{$K$-Means}
    \end{subfigure}
    \begin{subfigure}[t]{0.135\textwidth}
    \centering
    \includegraphics[width = \textwidth]{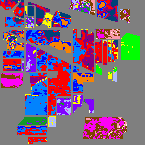} \hspace{0.02in}
    \vspace{-0.5cm}
    \caption{SC}
    \end{subfigure}
    \begin{subfigure}[t]{0.135\textwidth}
    \centering
    \includegraphics[width = \textwidth]{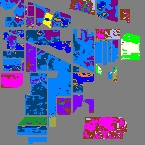} \hspace{0.02in}
    \vspace{-0.5cm}
    \caption{DPC}
    \end{subfigure}
    \begin{subfigure}[t]{0.135\textwidth}
    \centering
    \includegraphics[width = \textwidth]{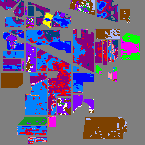} \hspace{0.02in}
    \vspace{-0.5cm}
    \caption{PGDPC}
    \end{subfigure}
    \begin{subfigure}[t]{0.135\textwidth}
    \centering
    \includegraphics[width = \textwidth]{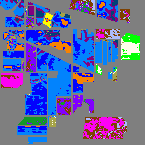} \hspace{0.02in}
    \vspace{-0.5cm}
    \caption{DL}
    \end{subfigure}
    \begin{subfigure}[t]{0.135\textwidth}
    \centering
    \includegraphics[width = \textwidth]{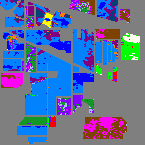} 
    \vspace{-0.5cm}
    \caption{D-VIC}
    \end{subfigure}
    
    \begin{subfigure}[t]{0.135\textwidth}
    \centering
    \includegraphics[width = \textwidth]{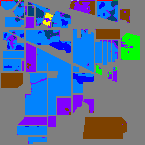} \hspace{0.02in}
    \vspace{-0.5cm}
    \caption{SC-I}
    \end{subfigure}
    \begin{subfigure}[t]{0.135\textwidth}
    \centering
    \includegraphics[width = \textwidth]{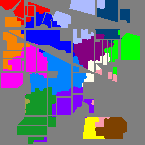} \hspace{0.02in}
    \vspace{-0.5cm}
    \caption{S-PGDPC}
    \end{subfigure}
    \begin{subfigure}[t]{0.135\textwidth}
    \centering
    \includegraphics[width = \textwidth]{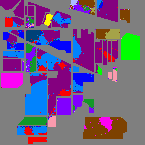} \hspace{0.02in}
    \vspace{-0.5cm}
    \caption{DLSS}
    \end{subfigure}
    \begin{subfigure}[t]{0.135\textwidth}
    \centering
    \includegraphics[width = \textwidth]{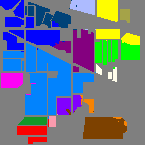} \hspace{0.02in}
    \vspace{-0.5cm}
    \caption{DSIRC}
    \end{subfigure}
    \begin{subfigure}[t]{0.135\textwidth}
    \centering
    \includegraphics[width = \textwidth]{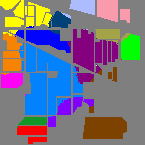} \hspace{0.02in}
    \vspace{-0.5cm}
    \caption{SRDL}
    \end{subfigure}
    \begin{subfigure}[t]{0.135\textwidth}
    \centering
    \includegraphics[width = \textwidth]{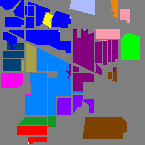} \hspace{0.02in}
    \vspace{-0.5cm}
    \caption{S$^2$DL}
    \end{subfigure}
    \begin{subfigure}[t]{0.135\textwidth}
    \centering
    \includegraphics[width = \textwidth]{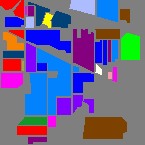} \hspace{0.02in}
    \vspace{-0.5cm}
    \caption{Ground Truth}
    \end{subfigure}
    
    \caption{Comparison of clusterings results of algorithms using spectral information (panels (b)-(g)), and algorithms using both spatial and spectral information (panels (h)-(l)) and S$^2$DL (panel (m)) on the Indian Pines dataset (panel (a)) with ground truth (panel (n)).}
    \label{fig:results_ip}
\end{figure*}

As shown in Figure \ref{fig:results_ip}, S$^2$DL exhibits outstanding performance on the Indian Pines dataset as well. Indeed, S$^2$DL achieved OA and $\kappa$ values slightly higher than those of SRDL---its nearest competitor---by approximately 1\% and AA values 4\% in  Table \ref{tab:datasets}. This improvement indicates S$^2$DL's enhanced accuracy in classifying diverse classes, including those with fewer labels. Furthermore, S$^2$DL delivers its efficient clustering performance with substantially lower runtime than SRDL. As discussed earlier, this significant reduction in computational cost is due to S$^2$DL's utilization of superpixelization. While SRDL utilizes a spatially-regularized graph, S$^2$DL integrates both superpixels and a spatially-regularized graph. This combination not only facilitates more efficient use of spatial information but also optimizes the computational process, leading to the observed runtime advantage.

\subsubsection{Hyperparameter Robustness Analysis} \label{sec: hyperparameter}

\begin{figure*}[ht]
    \centering
    
    \begin{subfigure}[t]{0.32\textwidth}
    \centering
    \includegraphics[width = \textwidth]{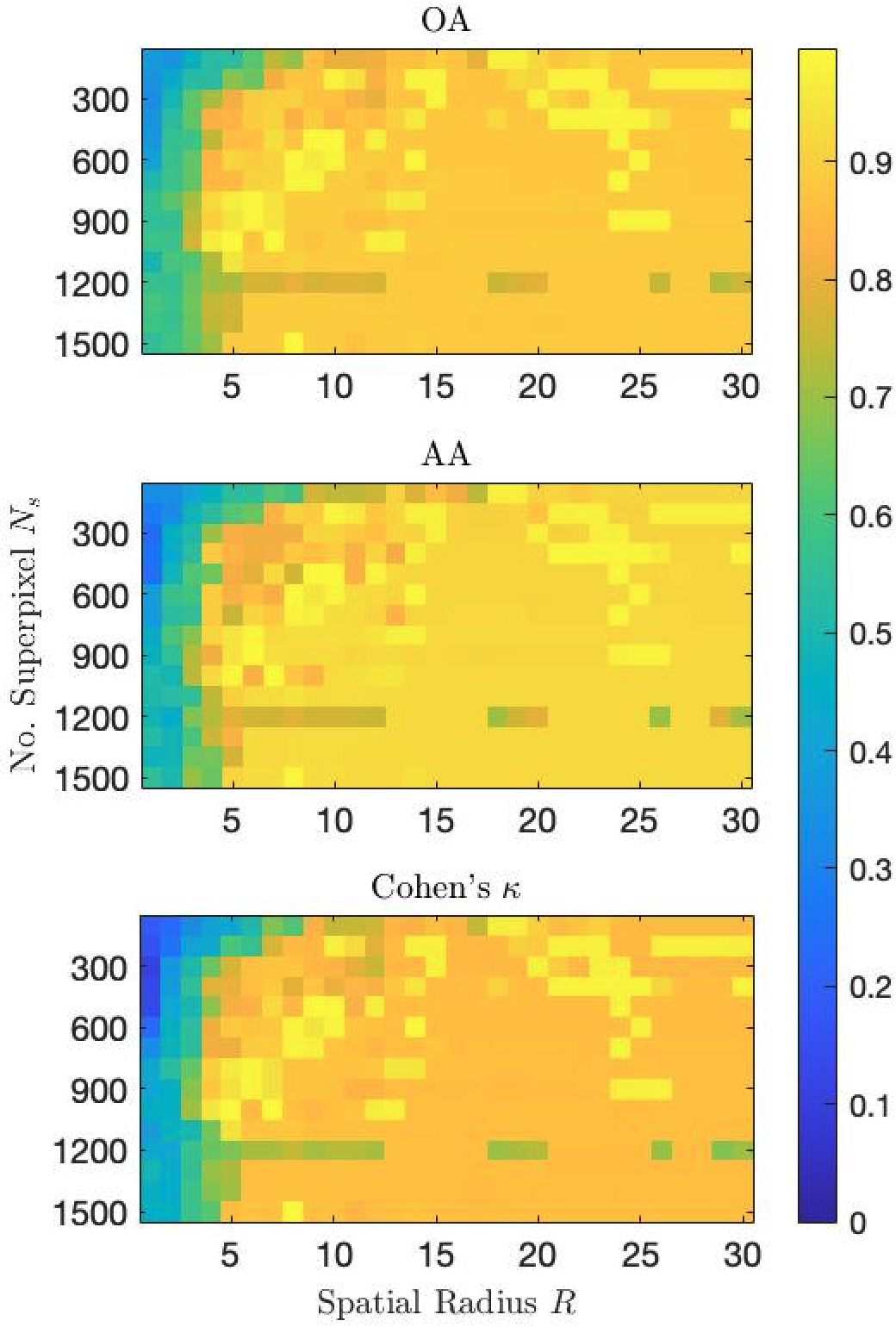}
    \vspace{-0.5cm}
    \caption{Salinas A, $k$=5}
    \end{subfigure}
    \begin{subfigure}[t]{0.32\textwidth}
    \centering
    \includegraphics[width = \textwidth]{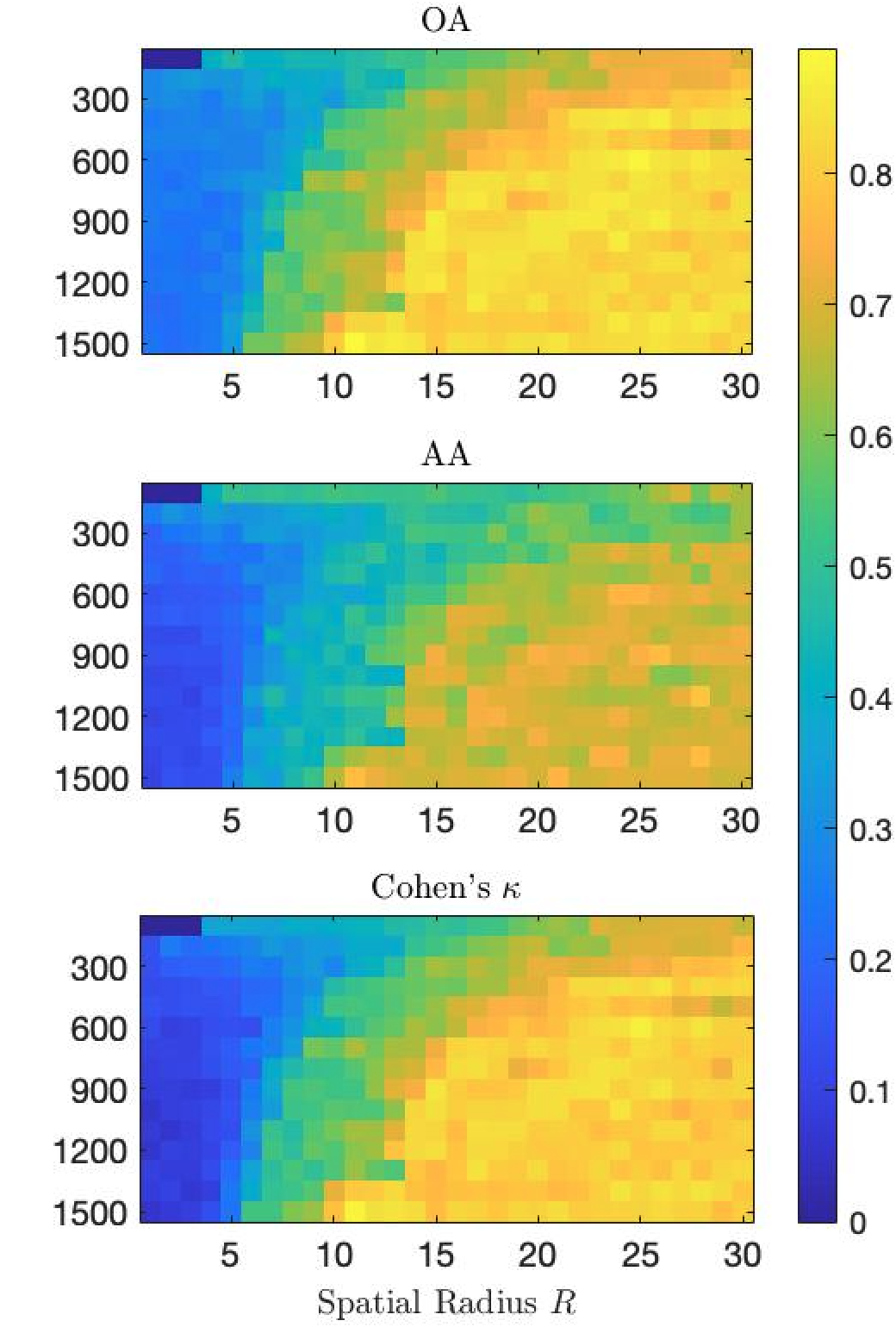}
    \vspace{-0.5cm}
    \caption{Salinas, $k$=3}
    \end{subfigure}
    \begin{subfigure}[t]{0.32\textwidth}
    \centering
    \includegraphics[width = \textwidth]{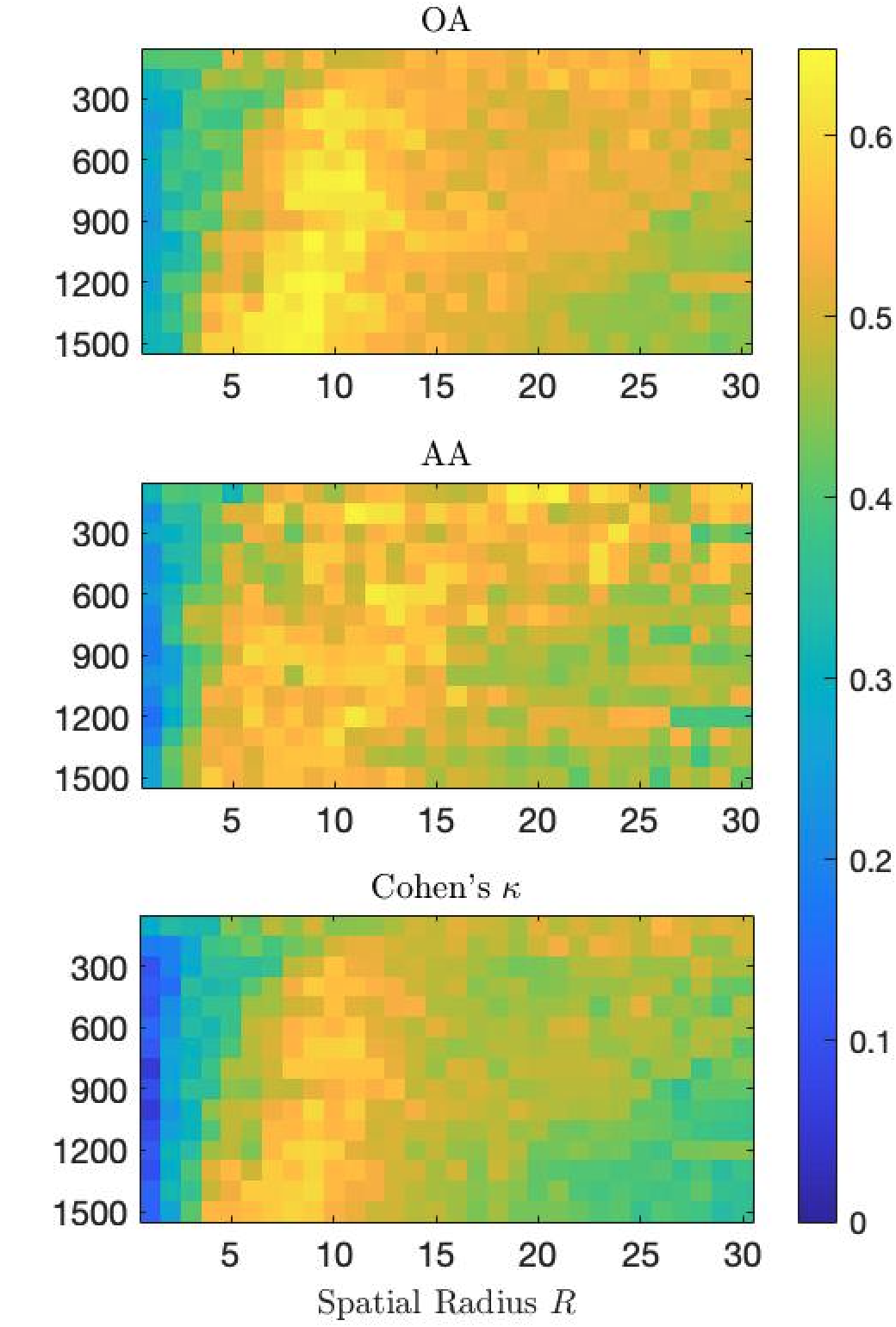}
    \vspace{-0.5cm}
    \caption{Indian Pines, $k$=5}
    \end{subfigure}
    
    \caption{Analysis of OA, AA, and $\kappa$ for three HSIs under varying spatial radii $R$ and numbers of superpixels $N_s$, with a fixed value of $k$. Rows in the figure represent the OA, AA, and $\kappa$ respectively, while columns correspond to the three different datasets. Each subplot within the figure depicts the accuracies achieved through various combinations of $R$ and $N_s$. The $x$-axis represents the spatial radius, while the $y$-axis denotes the number of superpixels.}
    \label{fig:robust_spr}
\end{figure*}

This section considers the robustness of S$^2$DL's clustering performance to the selection of hyperparameters. 
Figure~\ref{fig:robust_spr} illustrates the impact of the number of superpixels, denoted as $N_s$, and the spatial radius $R$ on clustering outcomes ($k$ held constant). An examination of the algorithm's performance on three HSIs suggests that a larger spatial radius $R$ is preferable when $N_s$ is relatively small. This strategy guarantees an adequate pixel count for the construction of a spatially-regularized graph. The Salinas dataset, characterized by its expansive size and homogeneous areas, requires a larger spatial radius compared to the other two datasets under consideration, and S$^2$DL's performance is fairly consistent when $R\in [10, 30]$ and $N_s\in [300,1500]$. On the other hand, the Indian Pines dataset, with its constrained spatial dimensions relative to the Salinas dataset, is more amenable to the selection of a smaller radius. As depicted in Figure~\ref{fig:robust_spr} (c), a similar result can be achieved when $R$ falls within $[5,15]$ and $N_s\in [300,1500]$. Regarding the Salinas A dataset, its size coupled with consistent spatial clusters make it suitable for a reduced number of superpixels. Notably, S$^2$DL can still yield high performance with $N_s<1000$. In general, there is an inverse relationship between the number of superpixels $N_s$ and the spatial radius $R$ across various datasets. The spatial dimensions and complexity of each dataset also influence the optimal selection of $N_s$ and $R$. Despite these variations, the S$^2$DL algorithm demonstrates robustness to changes in the hyperparameters $R$ and $N_s$, maintaining consistent performance across HSIs of different sizes and spatial complexities.

\begin{figure*}[ht]
    \centering
    
    \begin{subfigure}[t]{0.32\textwidth}
    \centering
    \includegraphics[width = \textwidth]{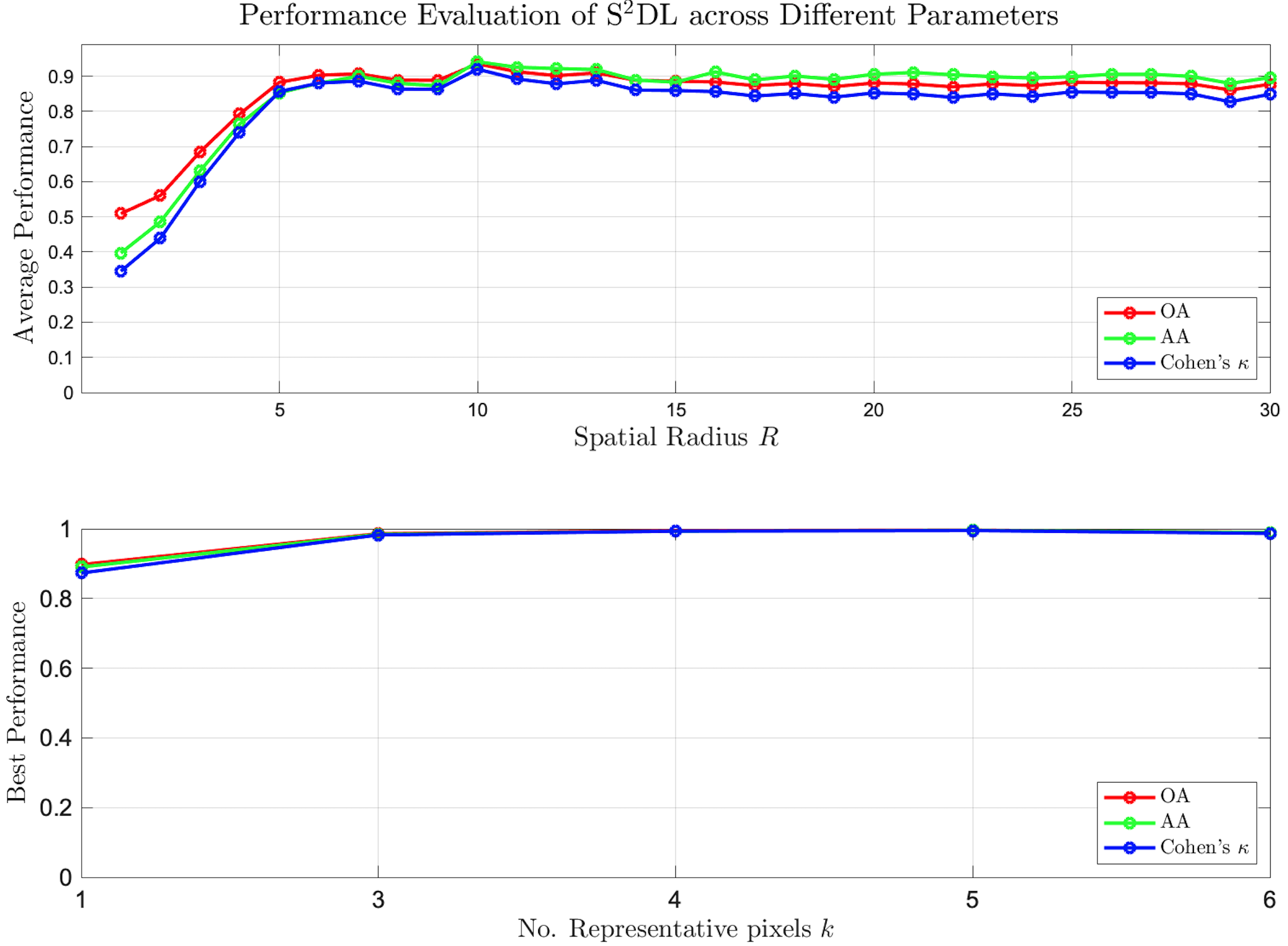}
    \vspace{-0.5cm}
    \caption{Salinas A, $N_s$=1000}
    \end{subfigure}
    \begin{subfigure}[t]{0.32\textwidth}
    \centering
    \includegraphics[width = \textwidth]{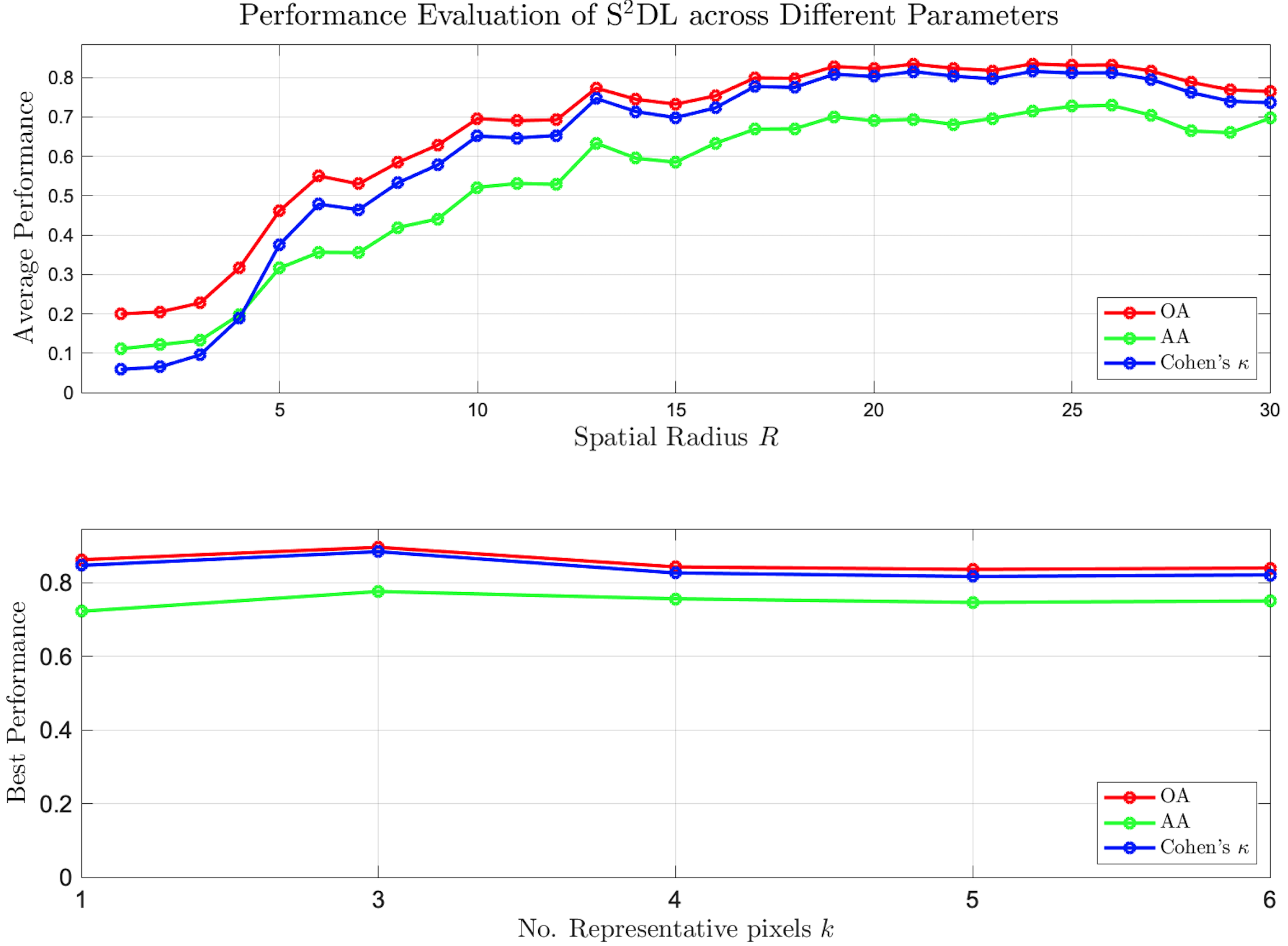}
    \vspace{-0.5cm}
    \caption{Salinas, $N_s$=1500}
    \end{subfigure}
    \begin{subfigure}[t]{0.32\textwidth}
    \centering
    \includegraphics[width = \textwidth]{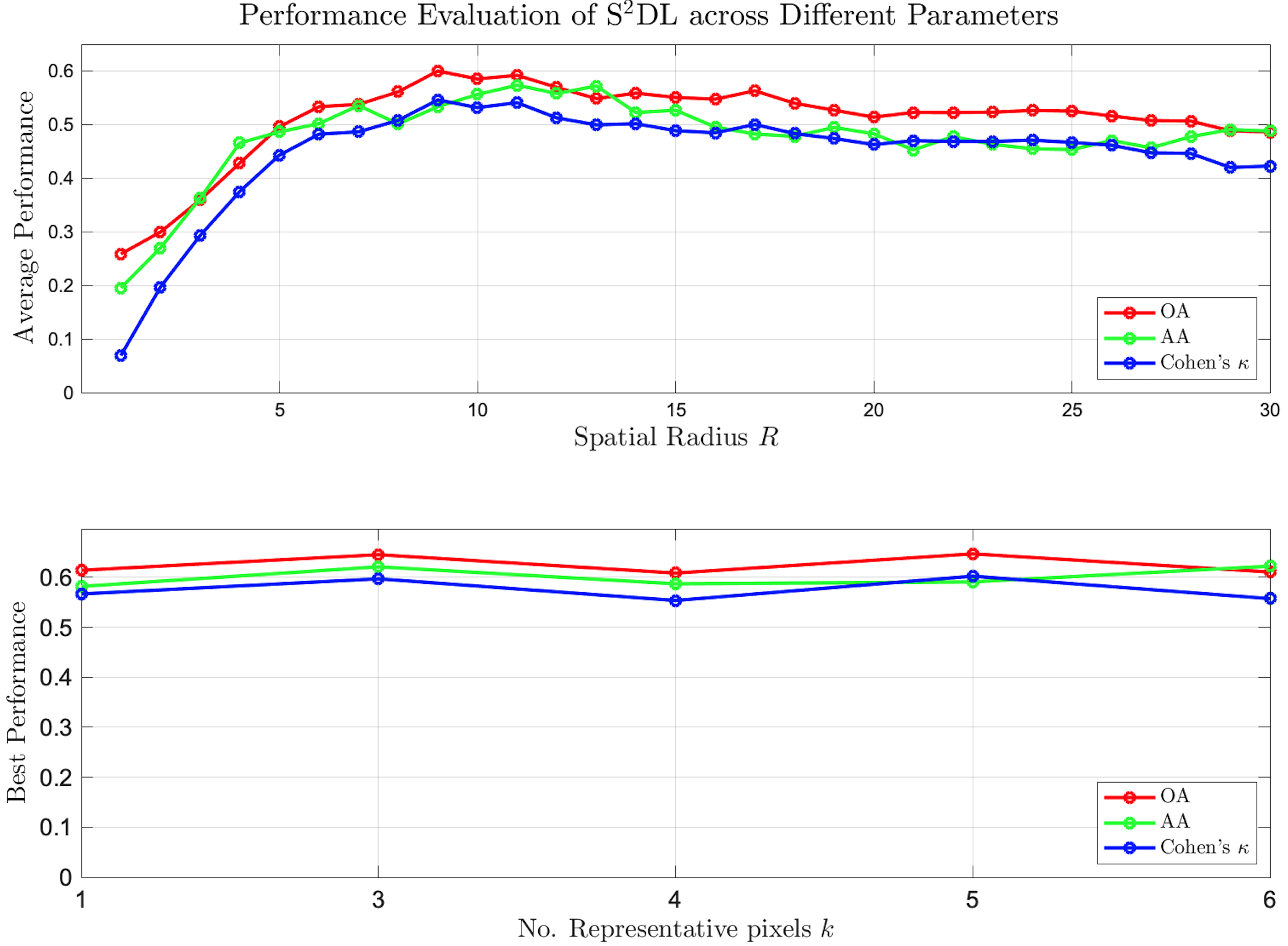}
    \vspace{-0.5cm}
    \caption{Indian Pines, $N_s$=1000}
    \end{subfigure}
    
    \caption{The average performance in relation to the spatial radius $R$ (first row) and the best performance concerning the number of representative pixels $k$ (second row). Each column corresponds to one of the datasets: Salinas A, Salinas, and Indian Pines, with a fixed number of superpixels.}
    \label{fig:robust_kr}
\end{figure*}

Figure~\ref{fig:robust_kr} quantifies the robustness of S$^2$DL to different spatial radii $R$ and number of representative pixels $k$. The top row of Figure~\ref{fig:robust_kr}  shows that, with an increase in spatial radius, the average performance across the three datasets initially experiences a swift uptick before eventually flattening. This pattern not only indicates robustness to changes in the spatial radius but also underscores the enhancement in results upon the integration of spatial information. For the Salinas A and Indian Pines datasets, the flattened phases both occur around $R=10$, while for the Salinas dataset, the flat phase arises near $R=20$. Similarly, Figure \ref{fig:robust_kr} indicates that S$^2$DL is highly robust to the selection of $k$---the number of representative pixels sampled from ERS superpixels---across the evaluated three datasets. We observe that S$^2$DL achieves highest performance on Salinas A for $k>3$ in Figure \ref{fig:robust_kr}. S$^2$DL achieves its peak performance on Salinas for $k$ small and has stable performance for $k>3$. Finally, although there is slight fluctuation in the performance of S$^2$DL on Indian Pines as $k$ increases, the variance in peak performance remains consistent within a narrow margin less than 5\%. This consistent trend across different values of $k$ demonstrates to the robustness of the proposed method for all three datasets under consideration.

\begin{figure*}
    \centering
    
    \begin{subfigure}[t]{0.32\textwidth}
    \centering
    \includegraphics[width = \textwidth]{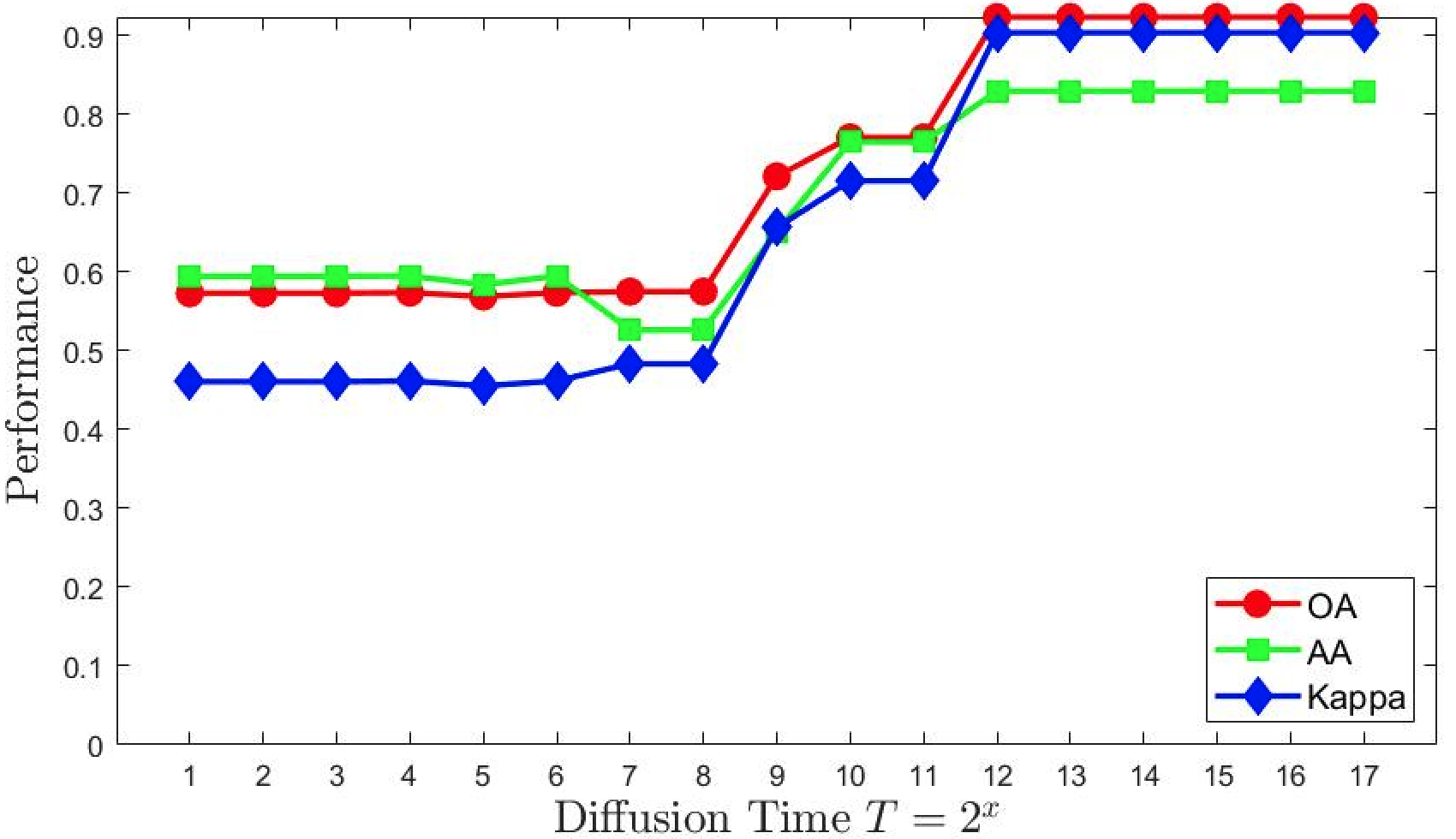}
    \vspace{-0.5cm}
    \caption{Salinas A}
    \end{subfigure}
    \begin{subfigure}[t]{0.32\textwidth}
    \centering
    \includegraphics[width = \textwidth]{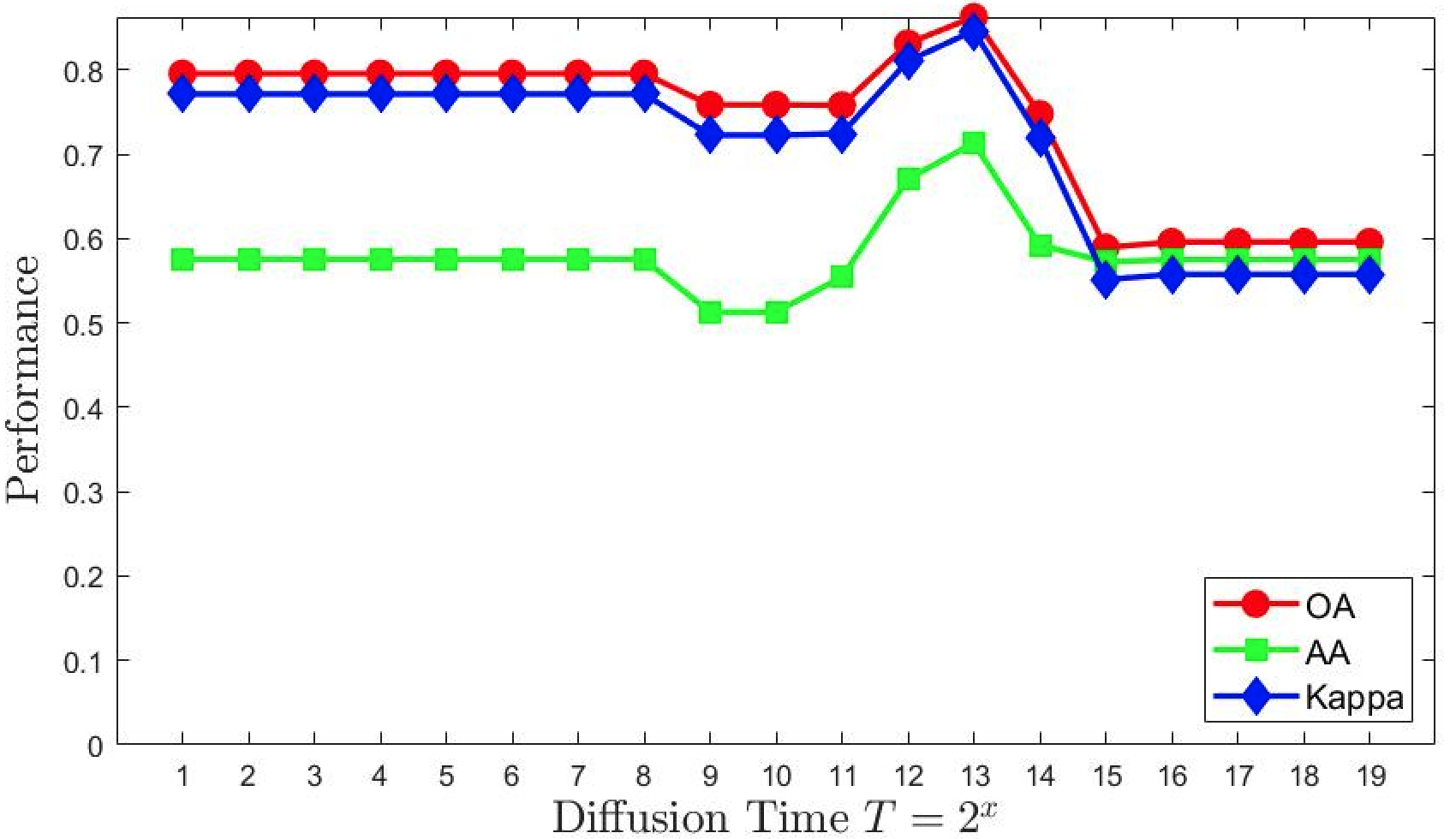}
    \vspace{-0.5cm}
    \caption{Salinas}
    \end{subfigure}
    \begin{subfigure}[t]{0.32\textwidth}
    \centering
    \includegraphics[width = \textwidth]{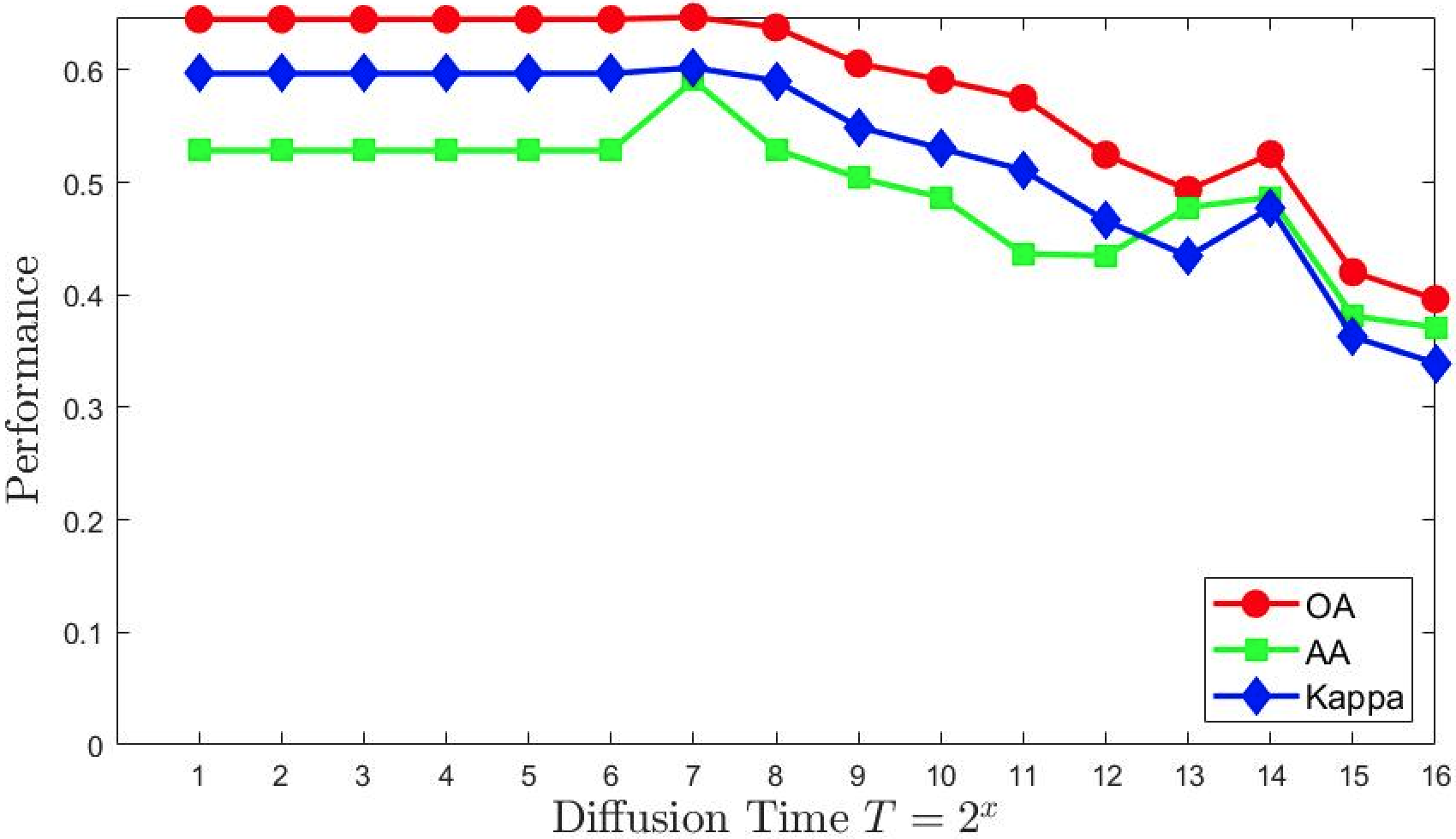}
    \vspace{-0.5cm}
    \caption{Indian Pines}
    \end{subfigure}
    
    \caption{Analysis of OA, AA, and $\kappa$ for S$^2$DL Across Various Diffusion Times: The $x$-axis represents the exponential scale of diffusion time (expressed as $2^{x}$), while the $y$-axis shows the corresponding values of OA, AA, and $\kappa$ as the time parameter $t$ varies across the diffusion process. This figure effectively illustrates the variation and stability of these performance metrics in response to changes in diffusion time.}
    \label{fig:robust_dt}
\end{figure*}

Finally, we analyzed the robustness of S$^2$DL to the selection of $t$: the diffusion time parameter used in diffusion distances. Specifically, we evaluated S$^2$DL at the optimal parameter set across a data-dependent exponential grid of $t$-values that captures the portion of the diffusion process during which cluster structure may be extracted; see Appendix \ref{app: hyperparameter} for more~\cite{polk2021multiscale, murphy2022multiscale, polk2023diffusion}.  In Figure \ref{fig:robust_dt}, it is evident that, for each dataset, there exists a wide window of diffusion time during which S$^2$DL achieves optimal performance. Notably, different algorithms required different diffusion time inputs for optimal clustering performance, likely due to differences in intrinsic geometric structure within the HSIs. That S$^2$DL is capable of recovering latent cluster structure during regions of time aligns with the literature in diffusion clustering~\cite{murphy2022multiscale, polk2021multiscale, coifman2006diffusion}, which has demonstrated that diffusion time is closely linked with the scale of discoverable cluster structure. Notably, this also indicates that S$^2$DL may be applied to the important problem of multiscale clustering by varying the diffusion time parameter $t$~\cite{murphy2022multiscale, polk2021multiscale, polk2022diffusion}. Regardless, the steady performance over extended periods of diffusion time indicates the algorithm's robustness to this parameter at the scale of interest.

\subsubsection{Ablation Study on Local Backbone}
\begin{figure*}
    \centering
    \includegraphics[width = 0.45\textwidth]{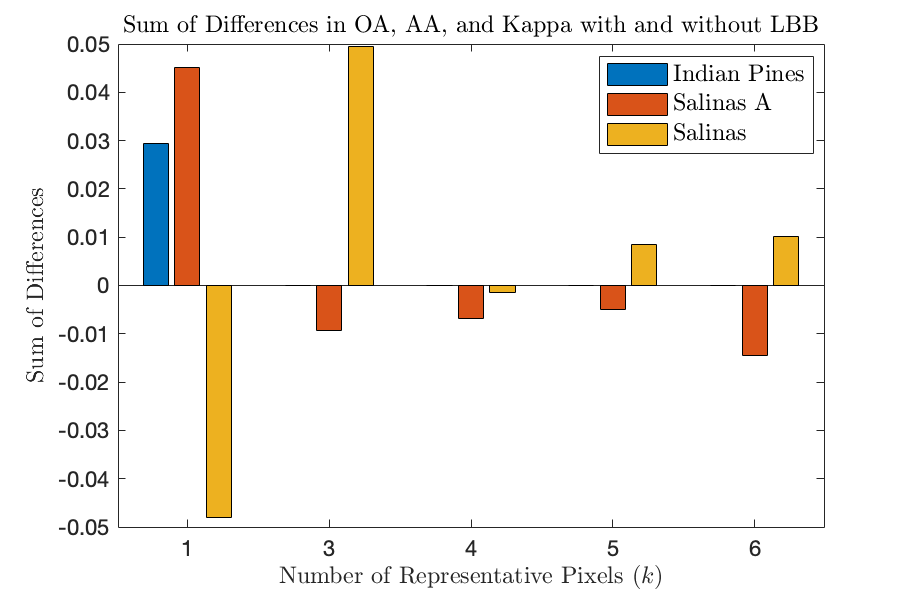}
    \caption{Performance Impact of Local Backbone Integration in S$^2$DL. This figure presents the summed performance differences across three metrics, calculated as the performance with LBB minus the performance without LBB.}
    \label{fig:robust_lbb}
\end{figure*}

This section provides an ablation study focused on the LBB~\cite{seyedi2019dynamic} within the S$^2$DL algorithm. Our objective is to empirically compare the performance of S$^2$DL with and without the implementation of LBB. The performance differences are quantitatively analyzed based on the three metrics analyzed throughout this section; see Figure \ref{fig:robust_lbb}.

Our analysis reveals distinct variations in performance across different datasets, elucidating the subtle role LBB plays in enhancing the algorithm's efficiency. For the Salinas dataset, LBB generally enhances performance metrics, although a notable exception occurs at $k=1$, where the limited pixel selection due to the dataset's size causes LBB to overemphasize pixel similarity. Conversely, S$^2$DL's performance on Indian Pines is uniform across $k$ for $k>1$, with or without the use of an LBB. Finally, on the Salinas A dataset, incorporating LBB leads to a slight reduction in over half of the metrics assessed, yet its  overall performance remains robust. This slight decline can be attributed to the smaller size of the Salinas A dataset and the subtle differences between classes; when 
$k$ is larger, the inclusion of more noisy pixels can diminish the reliability of LBB.

The efficacy of the LBB depends on the local homogeneity of the dataset and the quality of the modal pixels extracted. The S$^2$DL algorithm operates on a spatially-regularized graph, inherently conducting a localized search that solely focuses on the nearest neighbors of the identified modal pixels in the spatial-spectral domain. When the modal pixels identified are of high quality, the process of labeling their nearest neighbors significantly enhances the performance of the algorithm.

\subsection{Experiments on Mangrove Forests in Hong Kong}
\label{sec:hk}

The Mai Po Nature Reserve (MPNR), positioned at the entrance of the Shenzhen River in northwest Hong Kong and facing the Futian Nature Reserve in Shenzhen, spans the coordinates 113°59’E–114°03´E, 22°28´N–22°32´N. This reserve boasts a rich tapestry of ecosystems, including wetlands, freshwater ponds, inter-tidal mudflats, mangroves, reed beds, and fishponds, each fostering a diverse array of wildlife. Recognized for its ecological significance and unique location, MPNR was designated a restricted area in 1975 and subsequently declared a site of special scientific importance in 1976~\cite{wan2020gf}. At the heart of MPNR lies its expansive mangrove forests, covering approximately 319 hectares and recognized as Hong Kong's largest mangrove habitat~\cite{jia2014mapping}.

Originally, there were six native mangrove species within the Reserve, including Kandelia obovate (KO), Avicennia marina (AM), Aegiceras corniculatum (AC), Acanthus ilicifolius (AI), Bruguiera gymnorrhiza (BG) and Excoecaria agallocha (EA). The latter two species are rarer compared to their counterparts. Additionally, two exotic species, Sonneratia caseolaris and Sonneratia apetala, originating from the nearby Futian Nature Reserve, have been identified within MPNR. These are actively removed to mitigate their potential impact on the native mangrove population. Consequently, four dominant, well-studied species remain: KO, AM, AC, and AI. Specifically, KO and AI exhibit significant intraspecific variation, leading to their classification into sub-species: KO1, KO2, AI1, and AI2~\cite{wang2016textural}.

The hyperspectral data analyzed in this study were acquired using the Advanced Hyperspectral Imaging system aboard the Gaofen-5 Chinese satellite. This dataset encompasses 330 spectral bands, covering a 92$\times$72 pixel spatial region with a spatial resolution of 30~m. These bands include both the visible/near-infrared spectrum, with a spectral resolution of 5 nm, and the shortwave infrared spectrum, with a resolution of 10 nm~\cite{wan2020gf}. The study area includes not only the six primary mangrove classes but also classes such as mudflats and water bodies, collectively accounting for a total of 6624 pixels.

\begin{figure*}[t]
    \centering
    
    \begin{subfigure}[t]{0.4\textwidth}
    \centering
    \includegraphics[width = \textwidth]{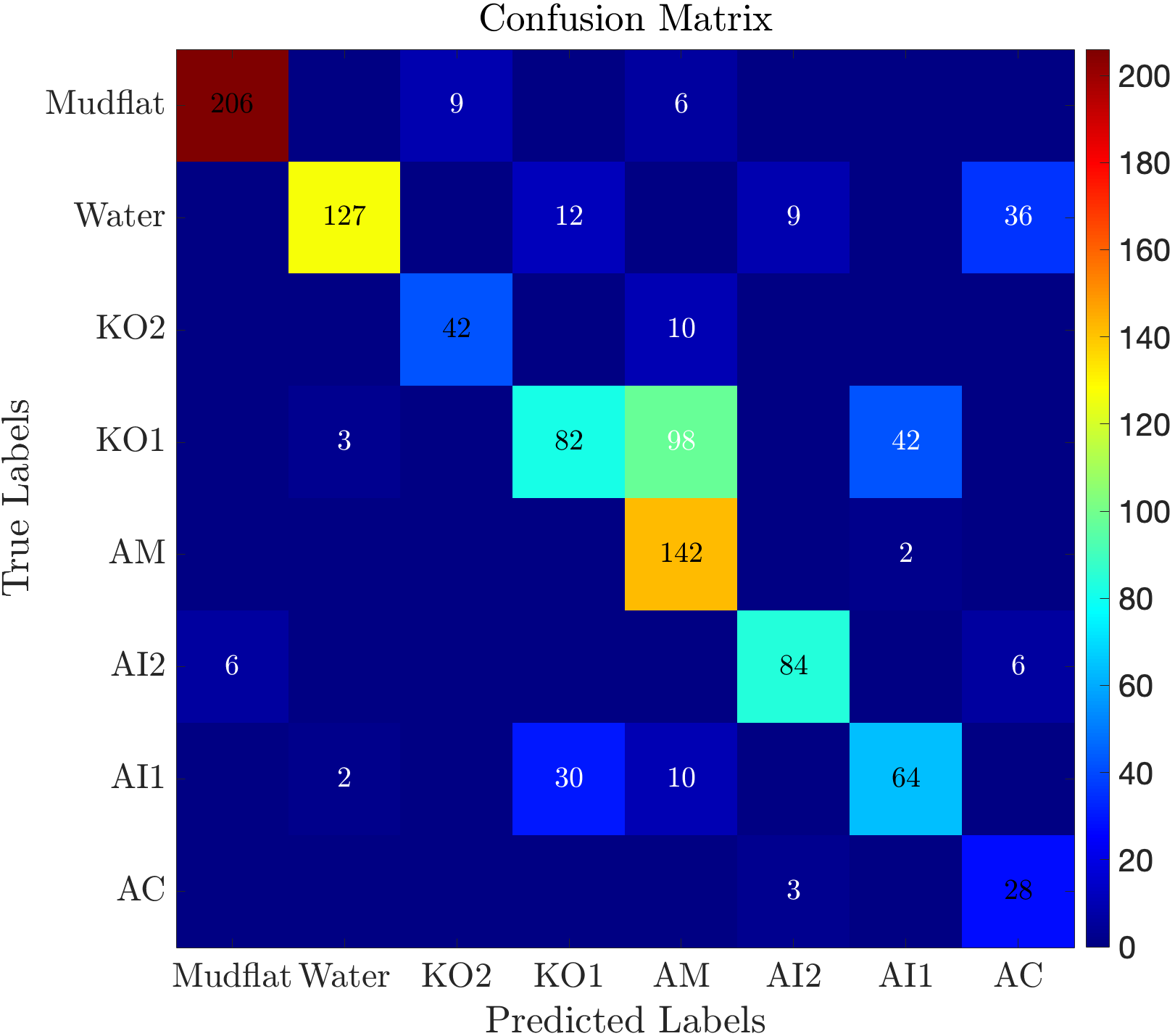} \hspace{0.02in}
    \vspace{-0.5cm}
    \caption{Confusion Matrix}
    \label{fig:conf}
    \end{subfigure}
    \begin{subfigure}[t]{0.37\textwidth}
    \centering
    \includegraphics[width = \textwidth]{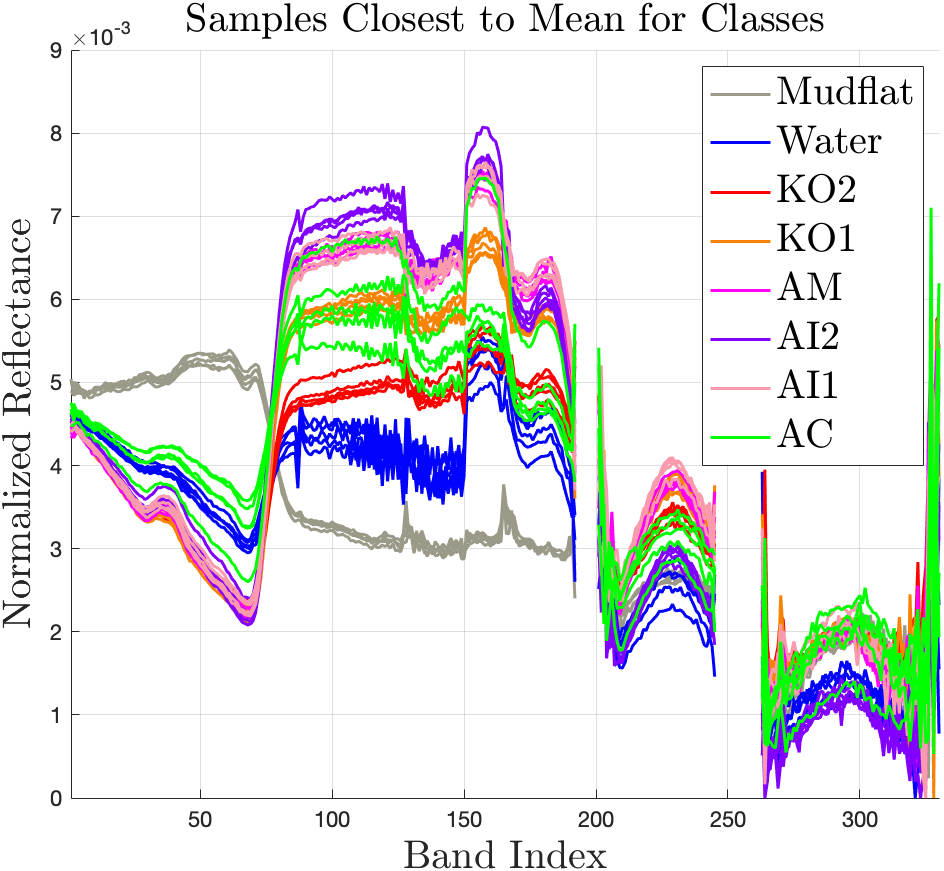}
    \vspace{-0.5cm}
    \caption{Spectral Signatures}
    \label{fig:ss}
    \end{subfigure}
    
    \caption{Confusion matrix for S$^2$DL clustering ($k=5$) and spectral signatures of samples from each class, colored by ground truth. Notably, different classes exhibit separation within spectral signatures in some spectral bands (but not all), making mangrove species mapping a challenging unsupervised remote sensing problem. }
    \label{fig:results_spectra}
\end{figure*}

\begin{figure*}[t]
    \centering
    
    \begin{subfigure}[t]{0.16\textwidth}
    \centering
    \includegraphics[width = \textwidth]{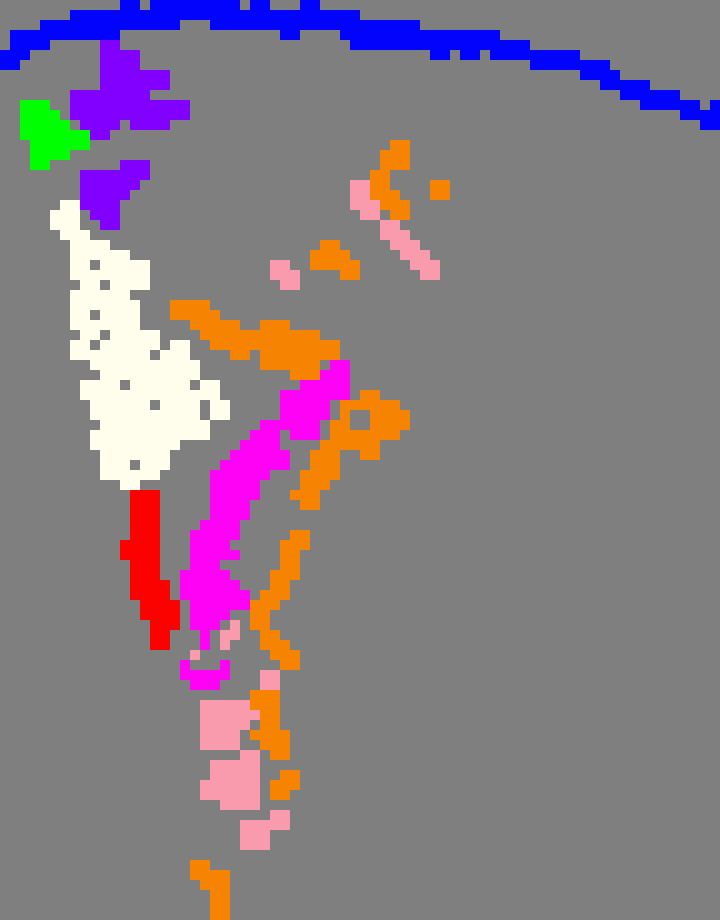} \hspace{0.02in}
    \vspace{-0.5cm}
    \caption{Ground Truth}
    \end{subfigure}
    \begin{subfigure}[t]{0.16\textwidth}
    \centering
    \includegraphics[width = \textwidth]{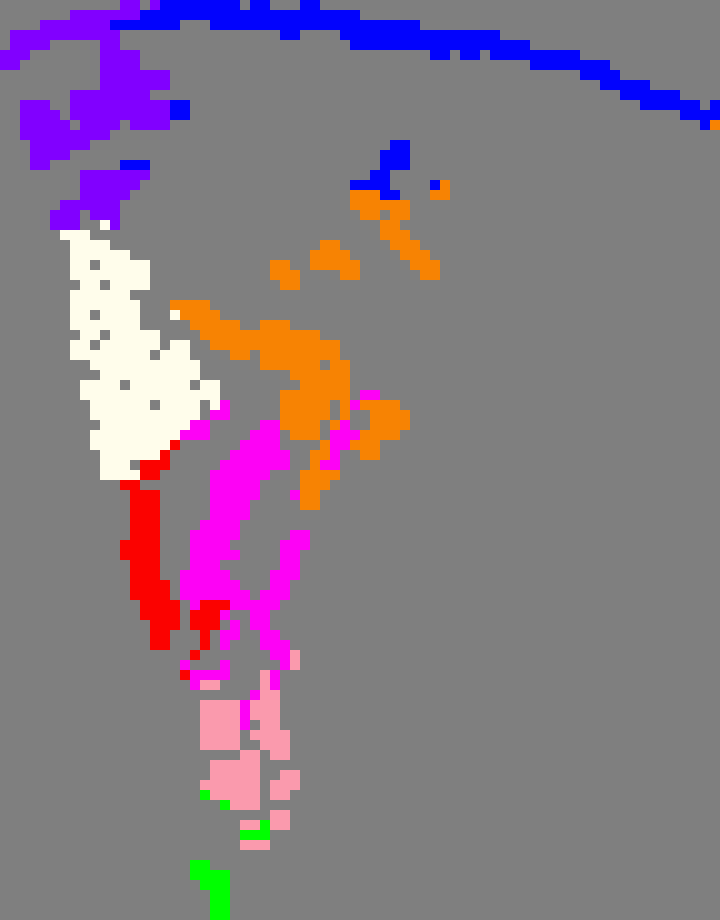} \hspace{0.02in}
    \vspace{-0.5cm}
    \caption{$k=1$}
    \end{subfigure}
    \begin{subfigure}[t]{0.16\textwidth}
    \centering
    \includegraphics[width = \textwidth]{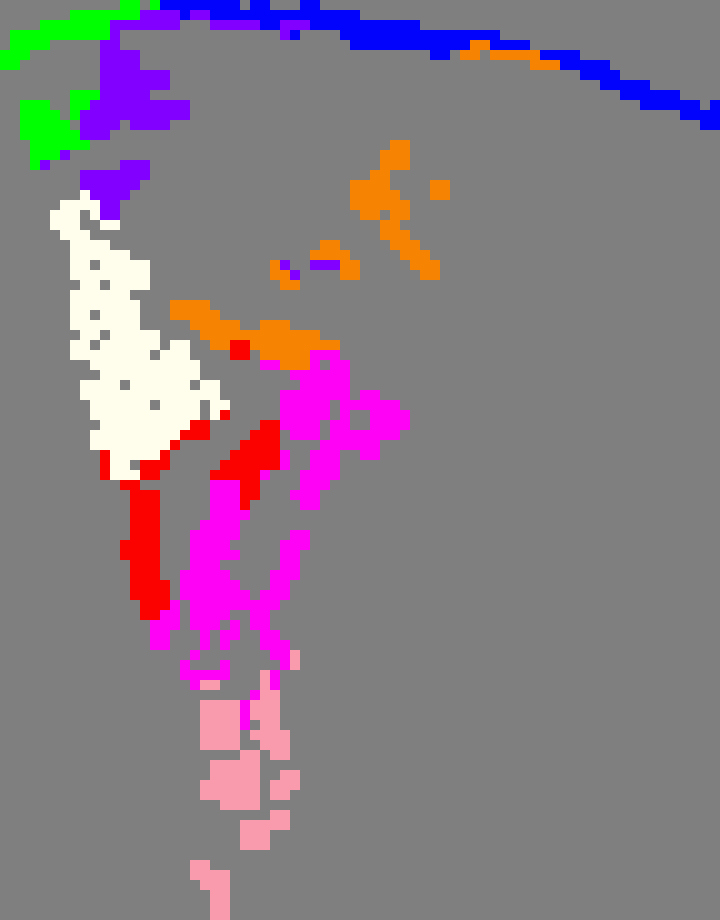} \hspace{0.02in}
    \vspace{-0.5cm}
    \caption{$k=3$}
    \end{subfigure}
    \begin{subfigure}[t]{0.16\textwidth}
    \centering
    \includegraphics[width = \textwidth]{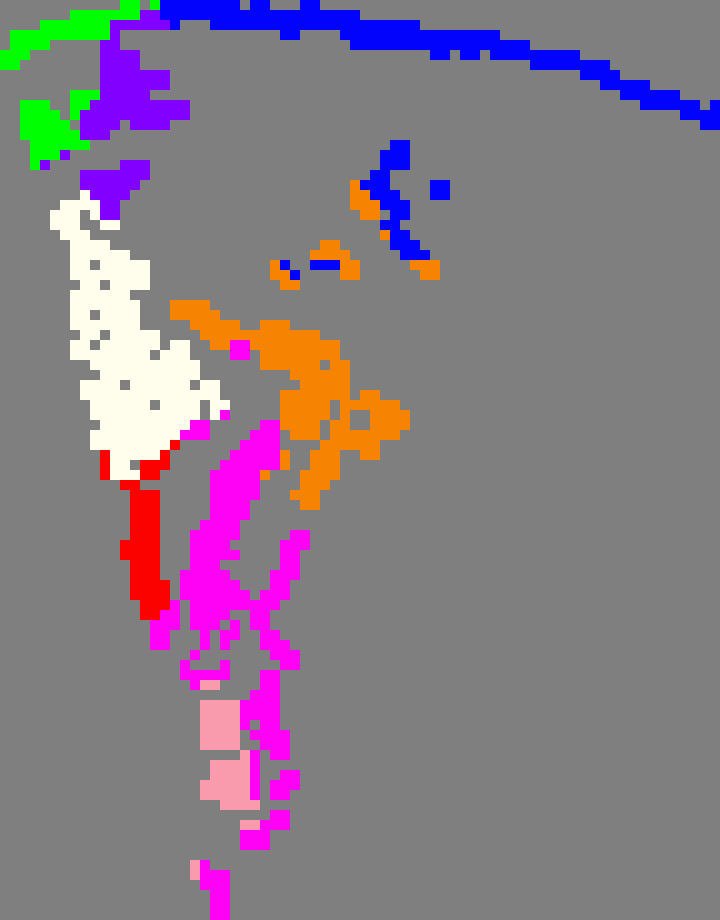} \hspace{0.02in}
    \vspace{-0.5cm}
    \caption{$k=4$}
    \end{subfigure}
    \begin{subfigure}[t]{0.16\textwidth}
    \centering
    \includegraphics[width = \textwidth]{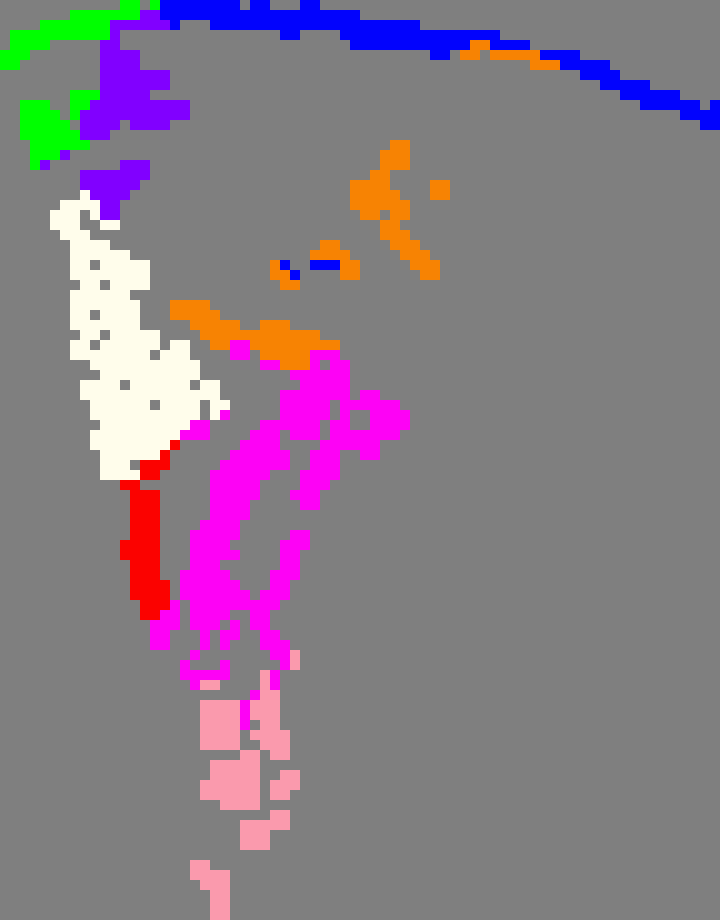} \hspace{0.02in}
    \vspace{-0.5cm}
    \caption{$k=5$}
    \end{subfigure}
    \begin{subfigure}[t]{0.16\textwidth}
    \centering
    \includegraphics[width = \textwidth]{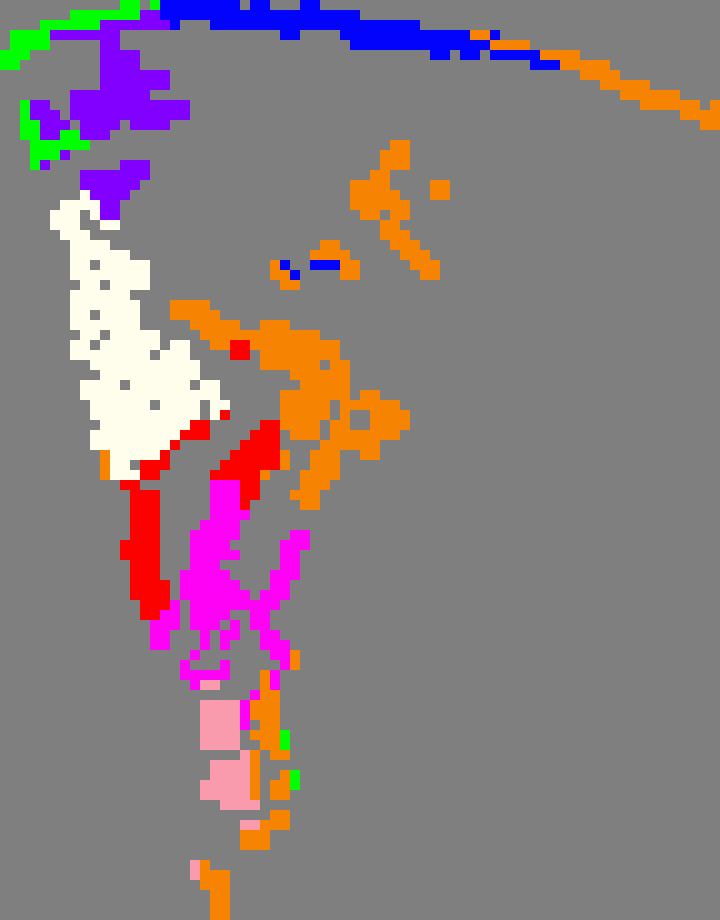} 
    \vspace{-0.5cm}
    \caption{$k=6$}
    \end{subfigure}
    
    \caption{Comparison of clustering results of S$^2$DL on the MPNR dataset with varying number of representative pixels per superpixel $k$. The color coding for the ground truth in (a) is as follows: grey for background, white for mudflat, blue for water, red for KO2, orange for KO1, pink for AM, purple for AI2, light rose for AI1, and green for AC.}
    \label{fig:results_mp}
\end{figure*}

The evaluation metrics employed for benchmark HSIs, including the introduction of producer's accuracy, are used in this study. This metric calculates the ratio of correctly classified pixels for a specific class to the total number of ground truth pixels for that class, thereby assessing class-specific performance. As depicted in Table~\ref{tab:mp}, our method demonstrates stable results across different values of the parameter $k$, with optimal performance observed at $k=5$, achieving an OA of 0.732, AA of 0.77, and $\kappa$ of 0.686. Notably, the best performance for half of the six mangrove species (spanning from column KO2 to AC) is achieved at $k=5$. S$^2$DL consistently outperforms other methods in OA, AA, and $\kappa$, with SRDL as its nearest competitor. S$^2$DL surpasses SRDL in half of the producer accuracy metrics across eight classes, while SRDL faces challenges in precisely identifying the AC and AM classes. In contrast, other methods struggle with accurately clustering classes like KO2, AI1, and AC due to sample imbalance, spectral similarities between classes, and diverse signatures within classes. S$^2$DL's success, therefore, highlights the benefit of using representative pixels from each superpixel to reduce the variability within spatial regions prior to cluster analysis.

Figure~\ref{fig:results_mp} presents the visualizations of S$^2$DL clustering outcomes for different values of $k$. The results are consistently high-performing across all settings, with most classes demonstrating optimal performance at $k=5$. Additionally, Figure~\ref{fig:conf} presents the confusion matrix for the S$^2$DL clustering with $k=5$, where most classes are effectively separated, with notable overlaps in predictions observed for classes such as KO1, AM, and AI1. As depicted in Figure~\ref{fig:ss}, which showcases the top 5 samples closest to the mean, AM and AI1 exhibit only subtle spectral differences. In contrast, KO1's spectral signature, which subtantially differs from AM and AI1 spectra, leads to the frequent misclassification by S$^2$DL of its upper-right and bottom pixels as AM and AI1, respectively, due to the spatial constraints in S$^2$DL. The high-quality unsupervised species mappings produced in this section underscore S$^2$DL's capability to deliver robust results in real-world forest environments.

\begin{table}[t]
\centering
\caption{Comparative Performance Analysis of S$^2$DL and Other Methods on the MPNR Dataset. The table presents the overall performance and producer's accuracy for varying $k$ values in S$^2$DL and compares it with other clustering methods. Best performances in each column for both S$^2$DL and other methods are highlighted in bold.}
\resizebox{0.8\textwidth}{!}{%
\begin{tabular}{|c|ccc|cccccccc|}
\hline
         & \multicolumn{3}{c|}{Overall Performance} & \multicolumn{8}{c|}{Producer's Accuracy} \\ \hline
S$^2$DL  & OA             & AA             & $\kappa$       & Mudflat        & Water          & KO2            & KO1            & AM             & AI2      & AI1     & AC      \\ \hline
$k=1$    & 0.715 & 0.669 & 0.664 & 0.887 & \textbf{0.794} & \textbf{1.000} & 0.538 & 0.667 & 0.917 & 0.547 & 0.000   \\
$k=3$    & 0.687 & 0.733 & 0.637 & 0.919 & 0.614 & 0.808 & 0.364 & 0.778 & 0.875 & \textbf{0.604} & \textbf{0.903}   \\
$k=4$    & 0.722 & 0.743 & 0.673 & 0.919 & 0.755 & 0.808 & 0.529 & 0.729 & 0.875 & 0.425 & \textbf{0.903}   \\
$k=5$    & \textbf{0.732} & \textbf{0.770} & \textbf{0.686} & \textbf{0.932} & 0.690 & 0.808 & 0.364 & \textbf{0.986} & 0.875 & \textbf{0.604} & \textbf{0.903} \\
$k=6$    & 0.706 & 0.688 & 0.652 & 0.919 & 0.522 & 0.808 & \textbf{0.796} & 0.521 & \textbf{0.938} & 0.425 & 0.581  \\ \hline
$K$-Means& 0.426 & 0.331 & 0.311 & 0.837 & 0.190 & 0     & 0.378 & \textbf{1.000}& 0     & 0     & \textbf{0.226}   \\
SC       & 0.533 & 0.502 & 0.452 & 0.805 & 0.147 & 0.635 & 0.480 & 0.882 & 0.844 & 0.104 & 0       \\
DPC      & 0.542 & 0.463 & 0.454 & 0.629 & 0.489 & 0     & 0.578 & 0.743 & 0.896 & 0.142 & \textbf{0.226}   \\
PGDPC    & 0.488 & 0.357 & 0.361 & 0.833 & 0.288 & 0     & \textbf{0.991}& 0     & 0.521 & 0     & \textbf{0.226}   \\
DL       & 0.542 & 0.463 & 0.454 & 0.629 & 0.489 & 0     & 0.578 & 0.743 & 0.896 & 0.142 & \textbf{0.226}   \\
D-VIC    & 0.566 & 0.466 & 0.476 & 0.701 & 0.446 & 0     & 0.693 & 0.826 & 0.823 & 0     & \textbf{0.226}   \\
SC-I     & 0.481 & 0.387 & 0.375 & 0.792 & 0.087 & 0     & 0.582 & 0.896 & 0.542 & 0     & 0.194   \\
S-PGDPC  & 0.651 & 0.542 & 0.586 & 0.891 & 0.571 & 0.019 & 0.529 & 0.785 & \textbf{0.958}& 0.585 & 0       \\
DLSS     & 0.592 & 0.490 & 0.512 & 0.805 & 0.467 & 0     & 0.653 & 0.806 & 0.927 & 0.038 & \textbf{0.226}   \\
DSIRC    & 0.549 & 0.466 & 0.456 & 0.842 & 0.005 & 0.365 & 0.680 & 0.806 & 0.938 & 0     & \textbf{0.226}   \\
SRDL     & \textbf{0.673}& \textbf{0.626}& \textbf{0.613}& \textbf{0.968}& \textbf{0.636}& \textbf{0.731}& 0.511 & 0.431 & \textbf{0.958}& \textbf{0.679}& 0.097   \\ \hline
\end{tabular}%
}
\label{tab:mp}
\end{table}

\section{Conclusion}\label{sec: conclusion}

This work introduces superpixel-based and spatially-regularized Diffusion Learning (S$^2$DL) for unsupervised HSI clustering. Given the high levels of noise and spectral variability often observed in the HSIs, algorithms that rely exclusively on spectral information fail to recover latent cluster structure or produce suboptimal material classifications~\cite{fauvel2012advances, zhai2021hyperspectral, plaza2009recent}. To mitigate these important challenges, S$^2$DL incorporates both spatial and spectral information, effectively processing HSIs that have consistent spatial patterns but also contain areas with noise or varied spectral characteristics. S$^2$DL demonstrates impressive clustering performance on both real-world benchmarking HSIs and the practical landscapes of the Mai Po Nature Reserve. This balanced performance highlights its robustness in standard scenarios and its adaptability to real-world environmental variations. Moreover, S$^2$DL is capable of achieving these high-quality clustering results at a fraction of the computational cost of related algorithms due to its reliance on superpixel segmentation prior to graph construction. Indeed, S$^2$DL's derived superpixels not only encapsulate localized spatial coherence within the image but also reduce the number of data to be analyzed, resulting in a computationally efficient clustering process and a robust utilization of spatial information~\cite{sellars2020superpixel, chen2023spectral}. Using a spatially regularized graph on this reduced dataset in a diffusion geometry-based clustering procedure enables S$^2$DL to efficiently leverage spatial information into a low-runtime and highly accurate clustering estimate~\cite{murphy2019spectral, polk2021multiscale}.

In future work, we aim to estimate the optimal number of superpixels by leveraging the intrinsic characteristics of datasets, such as size, spatial complexity, and resolution~\cite{fang2015classification, tu2018hyperspectral}. Since most common superpixel segmentation methods are designed primarily for RGB or grayscale images, they often fall short of fully extracting the abundant spatial and spectral information available in HSIs. Consequently, exploring and developing superpixel segmentation methods specifically tailored for HSIs will be worthwhile~\cite{sellars2020superpixel, barbato2022unsupervised}. Additionally, integrating feature extraction techniques into the S$^2$DL framework is a promising avenue, allowing us to utilize more effectively the rich spatial and spectral information within superpixels, thereby enhancing the overall performance and efficiency of the algorithm in clustering HSIs~\cite{superBF, fang2015spectral, cai2022superpixel, jiang2018superpca}. Furthermore, as referenced in Section \ref{sec: hyperparameter}, S$^2$DL is expected to be well-equipped to handle multiscale clustering problems by varying its diffusion time parameter. Moreover, by identifying the optimal clustering across scales through minimization of average variation of information~\cite{murphy2022multiscale, polk2021multiscale, meilua2007comparing}, we expect to be able to mitigate the dependence of S$^2$DL on diffusion time. While this may slightly reduce peak performance, it greatly enhances the practical applicability of the method. Lastly, pursuing the active extension of S$^2$DL, especially when a limited number of carefully selected labels are available depending on budget constraints, is a valuable direction for semi-supervised practical applications~\cite{maggioni2019active, murphy2020spatially, murphy2018unsupervised, polk2022active}.

\section*{Acknowledgments}
RHC was supported, in part by HKRGC GRF grants CityU1101120, CityU11309922, CRF grant C1013-21GF, and HKRGC-NSFC Grant N CityU214/19. JMM was supported, in part, by the National Science Foundation via grants DMS-2309519 and DMS-2318894. HZ was supported, in part, by HKRGC grants HKU27602020 and HKU17613022.

\section{Appendix:  Optimization of Hyperparameters} \label{app: hyperparameter}

This appendix details the process by which  hyperparameters were tuned in order to obtain the experimental results presented in Section \ref{sec: results}. Table \ref{tab:hyperparameters} provides a summary of the parameter grids for each algorithm. $K$-Means was implemented without the need for hyperparameter adjustments. For stochastic algorithms that require hyperparameter inputs (SC, D-VIC, and DSIRC), optimization was based on achieving the median sum of OA, AA, and $\kappa$ over 10 trials for each set of parameters in the specified hyperparameter grids.

\begin{table}[ht]
\centering
\caption{Hyperparameter ranges for algorithms, including $\mathcal{N}_1$ for exponential nearest neighbor sampling [10,900], $\mathcal{N}_2$ for spatially-regularized graphs [10,50], $\mathscr{D}$ for $\ell^2$-distances with 1000 nearest neighbors, $\mathscr{D}_1$ for SC-I distances, $\mathscr{D}_2$ for S-PGDPC Gaussian filter $\sigma$, $\mathscr{T}$ for diffusion time sampling, $\mathcal{B}$ for SC-I spatial-spectral information ratio, $\mathcal{R}$ for spatial regularization radii, $\mathcal{R}_1$ for DSIRC adaptive radius, and $\mathcal{S}$ for superpixel number range [100-1500]. '---' indicates no hyperparameter requirement.}
\resizebox{0.7\textwidth}{!}{%
\begin{tabular}{|c|c|c|c|c|c|}
\hline
 & Parameter 1 & Parameter 2 & Parameter 3 & Parameter 4 & Parameter 5 \\ \hline
$K$-Means  & --- & --- & --- & --- & --- \\ 
SC         & $k_n\in \mathcal{N}_1$ & --- & --- & --- & --- \\ 
DPC        & $k_n\in \mathcal{N}_1$ & $\sigma_0 \in \mathscr{D}$ & --- & --- & --- \\ 
PGDPC      & $k_n\in \mathcal{N}_1$ & --- & --- & --- & --- \\ 
DL         & $k_n\in \mathcal{N}_1$ & $\sigma_0 \in \mathscr{D}$ & $t \in \mathscr{T}$ & --- & --- \\ 
D-VIC      & $k_n\in \mathcal{N}_1$ & $\sigma_0 \in \mathscr{D}$ & $t \in \mathscr{T}$ & --- & --- \\ 
SC-I       & ---                    & $\sigma_1 \in \mathscr{D}_1$ & --- & $\beta\in \mathcal{B}$ & --- \\ 
S-PGDPC    & $k_n\in \mathcal{N}_1$ & $\sigma_2 \in \mathscr{D}_2$ & --- & --- & $N_s\in \mathcal{S}$ \\ 
DLSS       & $k_n\in \mathcal{N}_1$ & $\sigma_0 \in \mathscr{D}$ & $t \in \mathscr{T}$ & $R\in \mathcal{R}$ & --- \\ 
DSIRC      & $k_n\in \mathcal{N}_1$ & $\sigma_0 \in \mathscr{D}$ & $t \in \mathscr{T}$ & $R_1\in \mathcal{R}_1$ & --- \\ 
SRDL       & $k_n\in \mathcal{N}_2$ & $\sigma_0 \in \mathscr{D}$ & $t \in \mathscr{T}$ & $R\in \mathcal{R}$ & --- \\ 
S$^2$DL    & $k_n\in \mathcal{N}_2$ & $\sigma_0 \in \mathscr{D}$ & $t \in \mathscr{T}$ & $R\in \mathcal{R}$ & $N_s\in \mathcal{S}$ \\ \hline
\end{tabular}%
}
\label{tab:hyperparameters}
\end{table}

All graph-based algorithms in this study utilized adjacency matrices from sparse kNN graphs, with SC-I employing a full graph. For algorithms without spatial regularization, we used $\mathcal{N}_1$, an exponential sampling range of 10 to 900 for nearest neighbors. For those with spatial regularization, $\mathcal{N}_2$ was used, with values from 10 to 50 in increments of 10. DL, D-VIC, DLSS, DSIRC, SRDL, and S$^2$DL were executed at each $t$ within $\mathscr{T} = {0,1,2,2^2, \dots, 2^T}$, where $T=\lceil \log_2 [\log_{\lambda_2(\mathbf{P})}(\frac{2\times 10^{-5}}{\min(\pi)})]\rceil$. The process concludes at $t=2^T$ since, for $t\geq 2^T$, $\max_{x,y\in X}D_t(x,y) \leq 10^{-5}$~\cite{murphy2022multiscale}. The optimal time step $t$ from this range was selected for each dataset, maximizing the sum of OA, AA, and $\kappa$. Additionally, the KDE and $\sigma_0$ hyperparameters were uniformly applied across these algorithms. In our grid searches, $\sigma_0$ covered $\mathscr{D}$, which involved sampling $\ell^2$-distances between HSI pixels and their $k_n$ nearest neighbors. Additionally, $\sigma_1$ spanned $\mathscr{D}_1$ for SC-I, sampling $\ell^2$-distances between each data point and all others. For S-PGDPC, $\sigma_2$ was used as a parameter for Gaussian filtering, applied to blur the image prior to superpixel segmentation. For DLSS, SRDL, and S$^2$DL, the spatial radius $\mathcal{R}$ ranged from 1 to 30. DSIRC utilized $\mathcal{R}_1$ to automatically determine the radius of a spatially-adaptive window in various directions. SC-I employed $\mathcal{B}$ as the ratio parameter for balancing spatial and spectral information. The number of superpixels $N_s$ in $\mathcal{S}$ was set within a range of 100 to 1500 in increments of 100.

\printbibliography

\end{document}